\documentclass{article} % For LaTeX2e
\usepackage[final]{colm2026_conference} % <empty>, preprint, or final

\usepackage{microtype}
\usepackage{hyperref}
\usepackage{url}
\usepackage{booktabs}
\usepackage{comment}
\usepackage{algorithm}
\usepackage{algpseudocode}
\usepackage{lineno}

\author{\centerline{Quyet V. Do, Thinh Pham, Nguyen Nguyen, Sha Li, Pratibha Zunjare, Tu Vu} \\
\centerline{Virginia Tech, VA 24061, USA} \\
\centerline{\texttt{\{quyetdo,thinhphp,stephenallstar24,shal,pratibhaz,tuvu\}@vt.edu}}
}
\usepackage{enumitem}
\usepackage{amsmath}
\usepackage[utf8]{inputenc} % allow utf-8 input
\usepackage[T1]{fontenc}    % use 8-bit T1 fonts
\usepackage{hyperref}       % hyperlinks
\usepackage{url}            % simple URL typesetting
\usepackage{booktabs}       % professional-quality tables
\usepackage{amsfonts}       % blackboard math symbols
\usepackage{nicefrac}       % compact symbols for 1/2, etc.
\usepackage{microtype}      % microtypography
\usepackage{xspace}
\usepackage{soul}           % highlight
\sethlcolor{yellow}         % usage: \hl{text}
\usepackage{wasysym, dsfont}

\usepackage{array, multirow}
\usepackage{graphicx, subcaption}   % figures
\usepackage[dvipsnames]{xcolor} % colors
\usepackage{fontawesome} 
\usepackage{ragged2e} 

\usepackage{wrapfig}
\renewcommand{\arraystretch}{1.3}   % adjust row height

\usepackage{capt-of}    % or \usepackage{caption}
\usepackage{varwidth}   % for table-figure minipages
\usepackage{listings}   % print out prompt / data conveniently
\newif\ifcommentsoff
\newif\ifworkDivisionOff
\workDivisionOfftrue

\newcommand{\dataset}{$\pi^2$\xspace} 
\newcommand{\ourGpt}{\textsc{\dataset-20b}\xspace}
\newcommand{\ourQwen}{\textsc{\dataset-4b}\xspace}

\newcommand{\gptOssSmall}{\textsc{\small gpt-oss-20b}\xspace}
\newcommand{\qwenSmallInst}{\textsc{\small Qwen3-4B-Instruct-2507}\xspace}

\newcommand{\gptOssLarge}{\textsc{\small gpt-oss-120b}\xspace}
\newcommand{\qwenMedium}{\textsc{\small Qwen3.5-35B-A3B-FP8}\xspace}
\newcommand{\gptBest}{\textsc{\small GPT-5.4}\xspace} % too expensive to run

\newcommand{\geminiPro}{\textsc{\small Gemini-2.5-Pro}\xspace}
\newcommand{\geminiFlashLite}{\textsc{\small Gemini 3.1 Flash-Lite}\xspace}

\usepackage[most]{tcolorbox}
\tcbuselibrary{listings,breakable}

\newtcblisting{rawprompt}[1][]{
  breakable,
  listing only,
  colback=gray!5,
  colframe=gray!50,
  title=#1,
  listing options={
    basicstyle=\small\ttfamily,
    breaklines=true,
    columns=fullflexible,
    aboveskip=0pt,
    belowskip=0pt,
    breakindent=0pt,
    breakautoindent=false,
  }
}

\newcommand{\tabincell}[2]{\begin{tabular}{@{}#1@{}}#2\end{tabular}}
\newcommand{\textcolorregion}[2]{\textcolor{#1}{\parbox{\linewidth}{#2}}}
\definecolor{darkblue}{rgb}{0, 0, 0.5}
\hypersetup{colorlinks=true, citecolor=darkblue, linkcolor=darkblue, urlcolor=darkblue}

\title{\centering
\raisebox{-0.3em}{\includegraphics[height=1.8em]{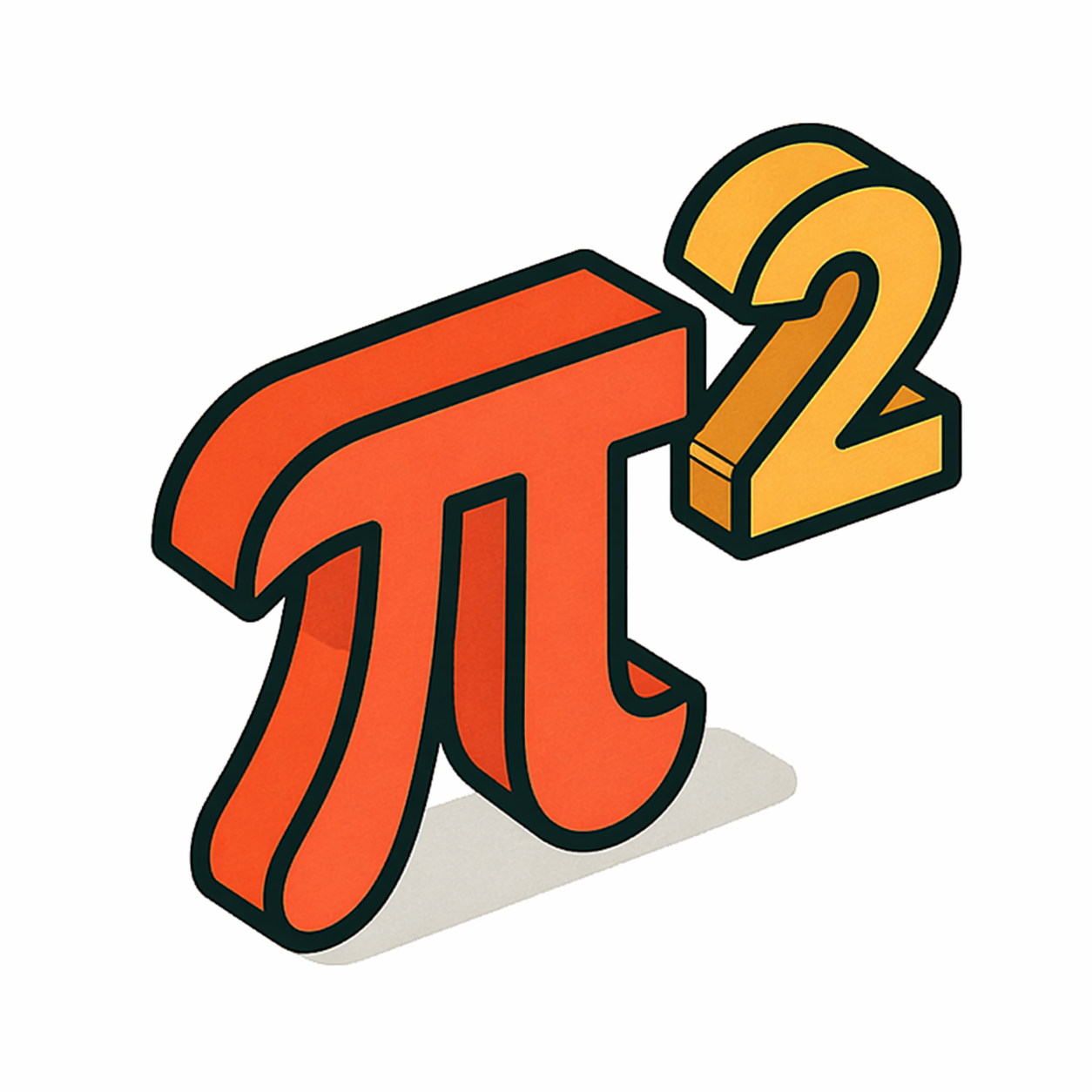}}%
  \hspace{0.3em}
Structure-Originated Reasoning Data Improves \\ Long-Context Reasoning Ability of Large Language Models
}

\begin{document}

\ifcolmsubmission
\linenumbers
\fi
\maketitle

% \note{Custom commands are defined in sections/custom-packages-commands.tex. LaTeX tips: https://saxon.me/blog/2025/latex-tips/.}

\begin{abstract}
We study a QA curation pipeline for improving long-context complex reasoning in large language models (LLMs). Our approach, $\pi^2$, constructs high-quality reasoning data through rigorous QA curation: 1) extracting and expanding tables from Wikipedia, 2) from the collected tables together with relevant metadata, generating complex reasoning questions whose answers are automatically determined and validated through dual-path code execution, 3) finally, back-translating chain-of-thoughts solutions grounded in realistic context.
Supervised fine-tuning with \textsc{\small{gpt-oss-20b}} and \textsc{\small{Qwen3-4B-Instruct-2507}} on $\pi^2$ yields consistent improvements across four long-context reasoning benchmarks and our alike $\pi^2$-Bench, with average absolute accuracy gains of +6.25\% and +3.37\% respectively. Through deeper analyses, we observe that reasoning style contributes little, while faithful reasoning patterns discovered by back translation and grounded realistic long context, as \dataset is designed for, are crucial for the improvement. Our code, data, and models are fully open-source at \url{https://github.com/vtpss/pi-squared}. % and attached to this submission.
\end {abstract}

% "Too Long; Didn't Read" (less than 20 words): Our complex reasoning data generated from Wikipedia tables and conditioned realistic documents improves LLM's long-context reasoning capability.
\section{Introduction}

\begin{quote}
\textit{"In all chaos there is a cosmos, in all disorder a secret order."} - Carl Jung
\end{quote}

%%% Motivation % DONE
Large language models (LLMs) have made rapid progress in handling long contexts, with recent systems supporting hundreds of thousands to millions of tokens ~\citep{openai2025gptoss, yang2025qwen3, openai2026gpt54, anthropic2026claude46, google2026gemini31}. This capability enables applications that require integrating information from multiple documents, such as open-domain question answering, document analysis, and web-scale search. However, improved context capacity does not translate to reasoning capacity, as current flagship models still struggle on complex reasoning over realistic long contexts ~\citep{zhu2024fanoutqa, bai2025longbench, chen2025browsecomp, pham2026sealqa, opsahl2026officeqa}.

%%% Problem and gap % DONE
A key bottleneck lies in the lack of verifiable supervision for complex reasoning over \textit{realistic long context in open-domain tasks}.
\citet{hf2025openr1} and \citet{guha2026openthoughts} represent a line of works that provide complex reasoning data with high-quality supervision, however, data domains are limited to natural science or coding.
In another line of works, \citet{yang2025longfaith, wang2025loongrl} and \citet{xie2026probing} ensure the verifiability of open-domain reasoning data, yet the contexts are synthetic or limited to tables, not realistic long documents.
Meanwhile, others were built on real open-domain documents and validated via code execution on concept graphs parsed from documents~\citep{yang2018hotpotqa, trivedi2022musique, chen2025longmit, shen2025qwenlong, sun2026cgmis}. Nonetheless, their complexity is limited to \textit{low branching-factor} reasoning patterns. % Like those question types in HotPotQA or MuSiQue.
This creates a gap: reliable supervision on complex reasoning is common in structured domains, but remains largely absent for realistic long contexts in open-domain tasks.

\newpage
\begin{wrapfigure}[15]{r}{0.52\textwidth}
\centering
\resizebox{\linewidth}{!}{%
    % print from drawio is more efficient than export.
    \includegraphics[width=1.0\textwidth]{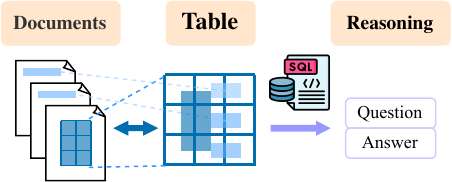}
}
\caption{Tables bridge realistic documents and complex reasoning in open-domain tasks. They \colorbox[HTML]{88B8FF}{represent data across sources}, while being \colorbox[HTML]{BBBBFF}{compatible with programmatic operations}.}
\end{wrapfigure}

\begin{figure*}[t]
% \vspace{-2em}
\centering
\resizebox{\linewidth}{!}{%
    % print from drawio is more efficient than export.
    \includegraphics[width=1.0\textwidth]{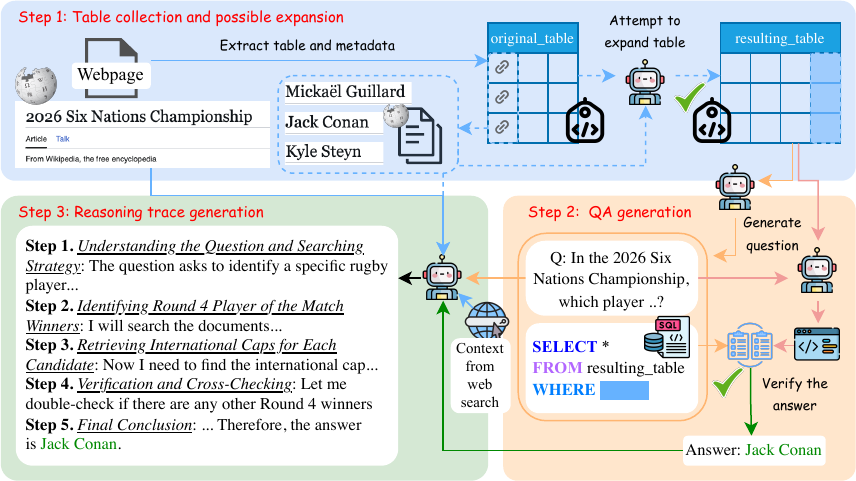}
}
\caption{\textbf{\dataset curation pipeline}. It first 1) \textit{collects tables from webpages and expands them} with information retrieved from linked documents \textit{if conditions are met} (indicated by dashed arrows and objects); then 2) from each collected table, \textit{generate a complex question} together with an executable SQL query by an LLM and \textit{verify the SQL-derived answer} with an independent LLM-generated Python solution, and finally 3) \textit{back-translates reasoning traces} grounded in real free-form context from the verified answer.}
\label{fig:data-pipeline}
\end{figure*}

%%% Core insight + example % DONE
We observe that structured representations, such as tables, is a natural medium that bridges complex reasoning and real long-context open-domain data, in which the data construction and verification are otherwise difficult.
Tables represent entities, attributes, and relations across sources, which can be losslessly altered by realistic documents. On the other hand, they remain compatible with operations such as filtering, aggregation, and sorting, which enables easy formation of \textit{high branching-factor} reasoning tasks and automatic verification of answers.
For example (\autoref{fig:data-pipeline}), consider the table "\texttt{\small Player of the Match awards}" in the "\texttt{\small 2026 Six Nations Championship}" Wikipedia page, where each row links to the player's page.
After augmenting this table with a new attribute \textsc{international\_caps} retrieved from linked pages, we can generate a complex reasoning question: \texttt{\small In the 2026 Six Nations Championship, which player has the highest number of international caps among Player of the Match winners in Round 4?}
Answering this question involves filtering by match round and selecting the maximum over caps (e.g., filtering rows where \textsc{round = 4}, then sorting by \textsc{international\_caps} and selecting the top entry), which can be verified exactly through an executable query.
Also, when the Wikipedia page and linked pages are provided instead of the augmented table, all the evidence is losslessly covered and distributed across multiple documents.

%%% Pipeline overview % DONE
Based on this observation, we propose \dataset (\textbf{Pi}peline for Generating \textbf{S}tructure-based \textbf{Qu}estion \textbf{A}nswering and \textbf{Re}asoning Data over Long \textbf{D}ocuments).
% a data curation pipeline that leverage structured representations to generate verifiable complex reasoning data over real long-context open-domain documents. % The info is too dense for comprehension.
The pipeline has three steps: 1) Collects tables from webpages and expands them with information retrieved from linked documents if conditions are met; 2) For each collected table, generate a complex question together with an executable SQL query by an LLM and verify the SQL-derived answer with an independent LLM-generated Python solution; 3) Finally, back-translates reasoning traces grounded in real free-form context from the verified answer. The generated data trains models to perform complex reasoning over realistic long documents in open-domain tasks.

%%% Results % DONE
We show that training on data generated with \dataset consistently improves long-context complex reasoning performance across multiple benchmarks, including in-domain LongSeal~\citep{pham2026sealqa}, OfficeQA~\citep{opsahl2026officeqa}, and our held-out evaluation set \dataset-Bench; as well as out-of-domain LongBench-v2~\citep{bai2025longbench} and Oolong-Synth~\citep{bertsch2025oolong}; with average gains of +6.25\% and +3.37\% for \gptOssSmall and \qwenSmallInst. Compared to the base models, finetuned models exhibit a stronger ability to reason over extended contexts, though it comes with a cost of overthinking. Through deeper analyses, we observe that reasoning style contributes little, while faithful reasoning patterns discovered by back translation and grounded realistic long context are crucial for the improvement. Finally, by the direct comparison with a previous long-context reasoning dataset, we show that \dataset closes the gap between previous works and real-world demands.

%%% Contributions
Our contributions are as follows:
\begin{itemize}[leftmargin=*,label=$\bullet$,noitemsep,partopsep=0pt,topsep=0pt,parsep=0pt]
    \item We propose \dataset, a QA curation pipeline that leverages tables as structured intermediate representations to generate verifiable, long-context, complex reasoning data across domains (\S\ref{sec:data}).
    \item We curate a dataset, also named \dataset, with a held-out benchmark, \dataset-Bench, for training and evaluating long-context complex reasoning in realistic settings (\S\ref{subsec:data-stats}, \S\ref{subsec:our-benchmark}).
    \item We show consistent improvements across multiple long-context reasoning benchmarks by finetuning models on \dataset, and that the improvements are attributed to faithful reasoning patterns discovered by back translation and grounded realistic long context (\S\ref{sec:experiments}).
    % \item \dataset enables effective self-distillation and data-efficient learning (\S\ref{sec:experiments}).
    % \item Finally, we release our dataset, benchmark, code, and finetuned models to facilitate reproducible research on long-context reasoning.
\end{itemize}

\section{Curation of \dataset}
\label{sec:data}

% Quyet + Nguyen: Bravo Thinh for a nice flow chart.
% It includes 1) Wikipedia table collection with possible expansion, 2) QA generation with code-based verification, and 3) Reasoning trace generation.

% Figure \ref{fig:data-pipeline} illustrates our data curation process, which consists of three main components: table collection and possible expansion, QA generation, and reasoning trace generation.

\subsection{Table Collection and Expansion}
\label{subsec:table}
% To ensure high data quality, diversity, realisticness, and factual grounding, we choose Wikipedia as the source for table collection. 
% We use the MediaWiki Action API to retrieve content and comprehensive metadata of Wikipedia pages.

We start with a diverse set of high-view and recent Wikipedia pages.\footnote{Accessed via the \texttt{pymediawiki} Python package} For each page, we extract tables and their associated metadata (e.g., caption, section, hatnote) by parsing the underlying HTML form.
Indeed, an extracted table represents itself as evidence in one document, while it is not certain if the table desirably represents evidence from multiple documents.
To actualize this property, we attempt to expand each extracted table with up to 3 new, contextually relevant columns whose content does not exist in the original Wikipedia source. \textit{In case a column in the extracted table contains reference links for all entries}, we use the infoboxes of those linked Wikipedia pages as external knowledge, then employ an LLM for the expansion. Post-expansion verification, including exact matching and LLM-as-a-judge, is executed to guarantee the information integrity and relevance of new columns.

\subsection{QA Generation}
\label{subsec:qa_gen}

For each resulting table (original one if the expansion failed) and its associated metadata, we prompt an LLM to generate a realistic complex question in natural language alongside its corresponding executable SQL query.
The question is designed to cover many columns of the table. If the table is expanded, the question should cover new columns as conditions or the target of QA. That resembles tasks which requires information aggregation from multiple real documents. Meanwhile, the question is expected to have a deterministic, unique, and short answer, as well as to be \textit{multi-hop reasoning with high branching factor}, i.e., to require multi-step linking, filtering, sorting, and/or numerical operations on attributes of multiple entities. The SQL query is then executed on the table, giving the candidate ground-truth answer.

We control the question's quality and the ground-truth answer's correctness as follows.
First, we prompt an LLM to judge the complexity, self-containedness,\footnote{\textit{Self-containedness} means "providing all necessary parameters, definitions, and background information within itself, requiring no external links, data, references, or prior knowledge to be understood"} and naturalness of the question in a Likert-5 scale. If the judge provides a low rating among three ratings, which indicates issues such as being overly simple, lack of clarity, or grammatical errors; the generation process is triggered to re-run (at most three times).
Second, we validate the SQL-derived answer with another independent computational path. We prompt an LLM to generate an independent Python-implemented solution of the question with the same tabular data as input. A generated QA sample is only retained if the answers from the SQL query and the Python script reach a consensus, determined by an LLM judge. More details of this step are shown in \autoref{app:dataset}.

% Samples of \dataset. All are taken from https://docs.google.com/spreadsheets/d/1NA9r52gzHD_7_xgTUMOf-pm7uH_e3Njb/edit?gid=1434114012#gid=1434114012

\begin{table*}[t]
    \small
    \centering
    \setlength{\tabcolsep}{4.5pt}     % narrow columns
    \renewcommand{\arraystretch}{1.1} % slightly taller rows
    \resizebox{\linewidth}{!}{%

%%%%%%%%%%
\begin{tabular}{p{0.58\linewidth}  p{0.12\linewidth} p{0.28\linewidth}}
\toprule
\multicolumn{1}{c}{\textbf{Question}} & \multicolumn{1}{c}{\textbf{Answer}} & \multicolumn{1}{c}{\textbf{Evidence}} \\

\midrule
Among the television roles held by Debbie Reynolds in her career, how many distinct character or guest roles did she perform specifically on shows broadcast by NBC? & 3 & Debbie Reynolds; Aloha Paradise; Alice (American TV series); etc. \\
\midrule
In the 2023 Cricket World Cup, among bowlers who use a Right-arm fast style, which player took the most wickets? & Mohammed Shami & 2023 Cricket World Cup; Mohammed Shami; etc. \\ % Adam Zampa;
\midrule
Based on the 2022 cobalt production and reserves data reported by the USGS, which country had the highest ratio of reserves to annual production? & Australia & Cobalt \\
\bottomrule
\end{tabular}
%%%%%%%%%%

    }
    % \vspace{0.0em}
    \caption{\textbf{Samples in \dataset}. The "Evidence" column shows the evidence Wikipedia page(s). The sample is originated from a table in the first listed page. Our samples cover diverse topics and complexity levels.}
    \label{tab:samples}
    % \vspace{-1em}
\end{table*}

\subsection{Reasoning Trace Generation}
\label{subsec:trace_gen}

For each validated question-answer pair, we first gather a corpus of evidence documents and possibly hard negative ones, which resemble real-world context. This corpus includes the Wikipedia pages used for table curation and ten relevant articles retrieved via automated web search with the question. To make the context manageable for long-context LLMs, we aggregate these sources and compact them to a maximum of 96K tokens, resulting in the \textit{compacted context}. We use BM25+~\citep{lv2011bm25plus} to rank the relevance to the question of all context passages, then iteratively discard the least relevant passage until the context length decreases to at most 96K. As location of evidence in Wikipedia pages is known, we ensure that all evidence,  remains in the compacted context. % including the original extracted table,

Finally, we prompt an LLM to articulate the \textit{reasoning steps} required to solve the question, given the compacted context and the verified answer. We note that the original extracted table is likely used as evidence in the reasoning traces, yet, retrieval steps from real context is required. For cases where the table is expanded, there must be further steps for evidence aggregation from multiple documents.

\subsection{\dataset}
\label{subsec:data-stats}

We started with 6,614 Wikipedia pages, collected 2,834 qualified tables and 1,172 validated QA samples with reasoning traces. We demonstrate 3 samples of \dataset in \autoref{tab:samples}. They cover diverse topics, question types, and complexity levels. Among the validated ones, \textit{369 samples are based on expanded tables}. For table expansion and QA curation, we used \qwenMedium. For reasoning traces generation, as we observed different reasoning "styles" of models from different families, we adaptively select the teacher model according to the target student model. \textit{We select \gptOssLarge/\qwenMedium if the student model is \gptOssSmall/\qwenSmallInst, respectively}. 
Overall, \dataset samples have context length ranging from 3.4K to 96K tokens with a median of 50.8K tokens, and reasoning trace length ranging from 500 to 10K tokens with a median of 1,480 tokens. \footnote{With respect to the \texttt{o200k\_base} encoding} 88.4\% of the questions requires at least 3 reasoning hops. More details on \dataset are shown in \autoref{app:dataset}.

\subsection{\dataset-Bench and Quality of \dataset}
\label{subsec:our-benchmark}

% Show some stats about human review for a high-quality benchmark.
% > "https://docs.google.com/spreadsheets/d/1NA9r52gzHD_7_xgTUMOf-pm7uH_e3Njb/edit?gid=1434114012#gid=1434114012"

To provide an additional measurement of long-context reasoning capabilities and to qualitatively assess the quality of our curated dataset, we held out and revised a challenging subset of \dataset to serve as a benchmark \dataset-Bench. In each sample of \dataset-Bench, models will be given the \textit{compacted context} and the question, and prompted to predict a short answer. Similar to previous works \citep{pham2026sealqa, chen2025browsecomp, zhu2024fanoutqa}, we used an LLM grader (with backbone \gptOssLarge) to judge the answer correctness. The grader template is provided in \autoref{app:prompt}. \textbf{For convenient notation, \dataset is referred to the training set} of the whole \dataset dataset by default, which is disjoint from the introduced \dataset-Bench.

\textbf{Sample selection and quality control of \dataset-Bench}: We deliberately prioritize difficulty and quality. In particular, we randomly selected 150 and 100 samples respectively from two pools: (1) samples originating from expanded tables with complexity ratings of 4 or 5, and (2) remaining samples with the highest complexity rating of 5. Each selected sample was reviewed by us to verify question quality and answer correctness against the Internet, including reference Wikipedia pages. That means, \textit{QA samples in \dataset-Bench are grounded to real-world information}. During this review process, 22 samples were discarded due to being difficult to revise without significant reconstruction. Among the remaining 228 samples, 209 required no revision, as the generated questions were already well-formed and complex as expected. % and only minor formatting adjustments to the answer format were needed. 
The other 19 samples required human intervention to either refine the question or correct the answer. Among 228 human-validated samples, we randomly select 50 samples for the development set. Remained 178 samples form the test set. % Deliberately do that, so that the test set is theoretically harder than the validation set.

\textbf{Notes on \dataset's quality}: A high yield was observed in this review process, where 209 out of 250 selected samples (83\%) required no revision. It shows the \textit{effectiveness of our data pipeline for the intended verifiable, long-context, complex QA}.

%%%%% Result on 4 benchmarks, excluding our benchmarks
% Better if having large models Gemini-2.5-Pro, GPT-5.4
% gpt-oss-120b + qwen3.5-35b-a3b-fp8: models used for pipeline generation
% gpt-oss-20b + qwen3-4b-instruct:
% - base, RLM, SFT (clean)
% best if also having PPO / GRPO

\begin{table*}[t]
    \small
    \centering
    \setlength{\tabcolsep}{4.5pt}     % narrow columns
    \renewcommand{\arraystretch}{1.1} % slightly taller rows
    % \resizebox{0.95\linewidth}{!}{%

%%%%%%%%%%
\begin{tabular}{lllllll}
\toprule
                    & \multicolumn{3}{c}{\textit{In-domain}}           & \multicolumn{2}{c}{\textit{Out-of-domain}}                \\
                    \cmidrule(lr){2-4}                                 \cmidrule(lr){5-6}
                    & LongSeal       & OfficeQA       & \dataset-Bench & LongBenchV2    & Oolong         & \textbf{Avg.}  \\

\midrule 
% \cmidrule(lr){2-6}
\geminiFlashLite    & \textbf{64.96} & \textbf{57.14} & \textbf{75.29} & 58.14          & \textbf{42.38} & \textbf{59.58} \\
\geminiPro          & 59.84          & 53.37          & 72.32          & 56.3           & 36.56          & 55.68          \\
\qwenMedium         & 58.5           & 55.74          & 76.97          & \textbf{58.22} & 39.76          & 57.84          \\
\gptOssLarge (low)  & 42.18          & 33.88          & 75             & 51.06          & 38.08          & 48.04          \\

\midrule 
\qwenSmallInst      & 29.14          & \textbf{15.12} & 54.68          & 36.09          & 30.51          & 33.11          \\
\ourQwen            & \textbf{37.01} & 14.53          & \textbf{55.35} & \textbf{41.12} & \textbf{34.41} & \textbf{36.48} \\

\midrule 
\gptOssSmall (low)  & 34.38          & 22.38          & 60.86          & 38.76          & 27.96          & 36.87          \\
\ourGpt (low)       & \textbf{39.69} & \textbf{30.40} & \textbf{68.01} & \textbf{44.52} & \textbf{32.98} & \textbf{43.12} \\

\midrule
\gptOssSmall (high) & 51.12          & 32.75          & 70.82          & 47.42          & 36.65          & 47.75           \\
\ourGpt (high)      & \textbf{61.09} & \textbf{41.91} & \textbf{73.93} & \textbf{50.73} & \textbf{37.59} & \textbf{53.05} \\

\bottomrule
\end{tabular}
%%%%%%%%%%

    % }
    % \vspace{0.0em}
    \caption{\textbf{\dataset significantly improves long-context reasoning ability of \qwenSmallInst and \gptOssSmall}. We evaluate our models and baselines on in-domain benchmarks LongSeal, OfficeQA, and our \dataset-Bench; as well as out-of-domain benchmarks LongBenchV2 and Oolong. For overall comparison, we report the weight-less average performance of models across benchmarks. \textbf{Bold-faced numbers} indicate the best among the group. We confirm that the average improvements brought by SFT on \dataset are significant at $\alpha = 0.01$ under t-test (i.e., with 99\% confidence level).
    % We note that we used \gptOssLarge/\qwenMedium as the model for reasoning back-translation if the student model was \gptOssSmall/\qwenSmallInst, respectively.
    }
    \label{tab:exp-main}
    % \vspace{-1em}
\end{table*}

\section{Experiments}
\label{sec:experiments}

\subsection{Methods and Baselines}

We perform supervised finetuning (SFT) with LoRA~\citep{hu2022lora} on \gptOssSmall and \qwenSmallInst with the 922 training samples of \dataset to obtain our models \textbf{\ourGpt} and \textbf{\ourQwen}, using hyperparameters outlined in \autoref{app:training}. The SFT process for each model took 1.5 hours on 4 NVIDIA H200 GPUs with PyTorch DDP.\footnote{Done with \texttt{unsloth} and \texttt{torchrun}} We report results with two reasoning effort configurations for \ourGpt (\texttt{low} and \texttt{high}), to assess the benefits of finetuning under different configurations.

We compare our finetuned models against the following baselines. \textbf{Base models}: The original \gptOssSmall and \qwenSmallInst without any finetuning on \dataset.
\textbf{Larger open-source models}: \gptOssLarge (\texttt{low} reasoning effort) and \qwenMedium, which represent a larger model scale and are models used for generating reasoning traces.
\textbf{Proprietary models}: \geminiPro and \geminiFlashLite,\footnote{Access via the \href{https://ai.google.dev/gemini-api/docs}{official Gemini API}} leading commercial LLMs. % More details on model evaluation setup are provided in \autoref{app:training}.

For open-source models, we run evaluations using vLLM~\citep{kwon2023vllm}.
\textbf{We report the average performance of the base models and our finetuned models across three runs}. Other evaluations, which serves as upper bounds, are run \textit{once}. We use models' default (i.e., recommended) hyper-parameters to maximize their performance.

\subsection{Benchmarks and Metrics}

We establish a comprehensive evaluation framework to assess our models and baselines across different dimensions of long-context reasoning. We consider both in-domain and out-of-distribution evaluations to understand the benefits of \dataset and the generalization capabilities of our models.

\paragraph{In-domain evaluation.}
Our in-domain suite contains benchmarks that have a large portion concerning realistic long-context QA tasks, with evidence grounded in tables in the context. % which is primarily targeted by our data curation pipeline
Along with our \textbf{\dataset-Bench}'s test set, the suite includes LongSeal and OfficeQA. \textbf{LongSeal}~\citep{pham2026sealqa} measures document-grounded QA in adversarial, noisy contexts. Here, each question's context contains the ground-truth document and 12 deceptive distractors. We checked the overlap between reference pages of LongSeal and \dataset, and confirmed no contamination. Meanwhile, \textbf{OfficeQA}~\citep{opsahl2026officeqa} straightforwardly evaluates end-to-end analytical reasoning on historical U.S. Treasury Bulletin documents. We follow the ``LLM with Oracle Parsed PDF Page(s)'' setup.

\paragraph{Out-of-domain evaluation.}
We evaluate our models on two benchmarks that differ in style or domain from our training data to probe distributional generalization. \textbf{LongBenchV2}~\citep{bai2025longbench} is a large-scale multiple-choice benchmark for reasoning over massive realistic contexts, covering various topics such as codebase QA, conversational, legal, financial reasoning, etc. Its ``Medium'' split (215 samples, context lengths 32k–128k) is used for evaluation. On the other hand, \textbf{Oolong} ~\citep{bertsch2025oolong} is an analytical reasoning benchmark requiring atomic text analysis and aggregation across long contexts. 400 Oolong-Synth test samples with 64k-token contexts are considered.

\paragraph{Metrics.}
We follow each benchmark's \textit{recommended metrics}. Notably, we use the provided LLM grader (with backbone \gptOssLarge) to report the accuracy on \dataset-Bench and LongSeal, while for others, we report scores from the \textbf{deterministic} graders. Because our focus is complex reasoning over long context rather than ultra-long-context tasks, inputs whose context exceeds 66\% of a 128K-token window are compacted with BM25+ in the same way as described in \S\ref{subsec:trace_gen}. % Additional experimental details are provided in \autoref{app:experiment}.

% DONE for the above if not otherwise marked.

\subsection{Results}
\label{sec:experiment-results}
% Qualitative sample from LongSeal

\begin{table}[t] % htbp
% \begin{wraptable}[51]{t}{0.4\textwidth}
    \small
    \centering
    \setlength{\tabcolsep}{4.5pt}     % narrow columns
    \renewcommand{\arraystretch}{1.1} % slightly taller rows
    % \resizebox{0.95\linewidth}{!}{%

%%%%%%%%%%
\begin{tabular}{l}
\toprule

\tabincell{p{0.95\linewidth}}{
\textbf{Question:} According to the WHO, in which year were the highest number of measles cases recorded worldwide? \\
\textbf{Correct Answer:} 1980 \\
\textbf{Context:}
% ... \\
% |          WHO-Region          |    1980   |    1990   |   2000  |   2010  |   2020  | \\
% |:----------------------------:|:---------:|:---------:|:-------:|:-------:|:-------:| \\
% ... \\
% | Worldwide                    | 3,852,242 | 1,325,074 | 853,479 | 343,806 | 159,073 | \\
% ... \\
% In 2019, the total number of cases worldwide climbed to 869,770. The number of cases reported for 2020 is lower compare to 2019. \\
% ... \\
(Omitted due to its length. A snapshot of evidence is shown in \autoref{fig:context-of-good-sample})
} \\

\midrule
\tabincell{p{0.95\linewidth}}{
\textbf{\gptOssSmall (low)}: \\
From the WHO data presented in the documents, \textcolor{BurntOrange}{the peak in worldwide measles cases occurred in 2019}, with a total of approximately 869,770 reported cases globally. This is higher than any preceding or following year reported in the sources (e.g., 2018 had 343,806 cases, and 2024 had 359,521 cases). Therefore, according to WHO statistics, \textcolor{BurntOrange}{2019} was the year with the highest number of measles cases worldwide.
} \\

\midrule
\tabincell{p{0.95\linewidth}}{
\textbf{\ourGpt (low)}: \\
The WHO’s own global measles surveillance data (published in Document 1) lists the total number of reported measles cases worldwide by year: 

\textcolorregion{blue}{
1980 – 3,852,242 cases, \quad 
1990 – 1,325,074 cases, \quad 
2000 – 853,479 cases \\
2010 – 343,806 cases, \quad 
\;\;\,2019 – 869,770 cases, \quad 
\;\;\,2020 – 159,073 cases
}
Among these figures, \textcolor{blue}{the largest reported number is the 3,852,242 cases recorded for the year 1980}. No other year shown in the official WHO data exceeds this count, and [\textit{31 words omitted for brevity}]. Therefore, according to the WHO’s own records, the highest number of measles cases worldwide occurred in \textcolor{blue}{1980}.
} \\

\bottomrule
\end{tabular}
%%%%%%%%%%

    % }
    % \vspace{0.0em}
    \caption{\textbf{\dataset equips models with stronger long-context retrieval and reasoning}. We show the response of \gptOssSmall and \ourGpt (with \texttt{low} reasoning effort) on a sample of LongSeal. The base model \textcolor{BurntOrange}{fail to reason and retrieve} the correct piece of evidence, as it mistakes the phrase "climbed to 869,770 in 2019" as the signal for the correct answer, thus missing out other evidence. On the other hand, \ourGpt \textcolor{blue}{successfully identifies all pieces of evidence and reason thoroughly} through them to derive the correct answer. That is a core capability targeted by our data curation pipeline.}
    \label{tab:good-qualitative-example}
    % \vspace{-1em}
\end{table}
% \end{wraptable}

\autoref{tab:exp-main} presents our main results, demonstrating that \dataset significantly improves the long-context reasoning capability of base models across benchmarks. Below are our key findings.

\textbf{Consistent improvements across models and benchmarks.} % DONE
Supervised finetuning on \dataset yields improvement for models in different ways.
\ourGpt significantly outperforms the base \gptOssSmall on the in-domain benchmarks (LongSeal, OfficeQA, and \dataset-Bench), with improvements of 5.31, 8.02, and 7.15 points under \texttt{low} reasoning effort, and even larger gains of 9.97, 9.16, and 3.11 points under \texttt{high} reasoning effort. Regarding the out-of-domain benchmarks, the improvement is, though less pronounced, still remarkable. While \ourGpt (high) slightly improves on Oolong and  LongBenchV2, \ourGpt (low) has a 5.03-point improvement on Oolong and a 5.77-point improvement on LongBenchV2.
Meanwhile, \ourQwen also shows a large improvement on LongSeal but not on OfficeQA and \dataset-Bench among in-domain benchmarks. We attribute this phenomenon to the fact that \qwenSmallInst is a smaller model and has a weaker reasoning capability than \gptOssSmall, thus benefiting less from the reasoning patterns in \dataset. This phenomenon has been observed in previous works~\citet{li-etal-2025-small-models, ye2025limo}.
Nonetheless, \ourQwen interestingly outperforms the base model 5.03 and 3.90 points on LongBenchV2 and Oolong, respectively.
Although each model absorbs a different skill set from \dataset, the overall improvements are consistent, indicating that the \textit{benefits of \dataset are not limited to a specific data distribution or pre-trained model.}

\textbf{Narrowing down the gap with proprietary and larger models.} % DONE, no need to modify
While proprietary models \geminiFlashLite, \geminiPro, and larger open-source models such as \qwenMedium maintain an overall lead, \ourGpt (high) substantially narrows the gap. Notably, on complex reasoning benchmarks such as LongSeal (61.09 vs. 58.50 for \qwenMedium) and OfficeQA (41.91 vs. 55.74 for \qwenMedium), the gap is largely reduced, demonstrating that high-quality training data can partially compensate for model scale.

% Qualitative and output-length
\textbf{Analyses.}
We present two samples from the LongSeal benchmark in \autoref{tab:good-qualitative-example} and \autoref{tab:not-good-qualitative-example} to qualitatively illustrate the impact of \dataset on model's reasoning capability. \ourGpt demonstrates skills for comprehensive evidence retrieval and complex reasoning, thus not making a mistake like the base model. That is a core capability targeted by our data curation pipeline. Nonetheless, as not primarily designed for, \dataset does not help models overcome reporting bias. Furthermore, the stronger reasoning capability of our finetuned models is accompanied by a cost of overthinking, as the output length of \ourGpt is significantly longer than the base model. We provide more analysis on this phenomenon in Appendix \ref{app:length-analysis}.

\section{Ablation Study}
\label{sec:ablation}
% Qualitative sample from LongSeal

\begin{table}[t] % htbp
% \begin{wraptable}[51]{t}{0.4\textwidth}
    \small
    \centering
    \setlength{\tabcolsep}{4.5pt}     % narrow columns
    \renewcommand{\arraystretch}{1.1} % slightly taller rows
    % \resizebox{0.95\linewidth}{!}{%

%%%%%%%%%%
\begin{tabular}{l}
\toprule

\tabincell{p{0.95\linewidth}}{
\textbf{Question:} Who holds the all-time record at the Grammys for the most wins in the album of the year category? \\
\textbf{Correct Answer:} Serban Ghenea \\
\textbf{Context:} (Only show the most important evidence. Omitted remaining text for brevity.) \\
\textit{[In Document 1: "Grammy Award for Album of the Year" Wikipedia page, as a document in the context]} \\
... \textcolor{blue}{Serban Ghenea} (as engineer/mixer) is the most frequent winner in this category with five awards. Taylor Swift (as performer); John Hanes (as engineer/mixer); and Tom Coyne and Randy Merrill (as mastering engineers) have won the award four times ...
} \\

\midrule
\tabincell{p{0.95\linewidth}}{
\textbf{\gptOssSmall (low)}: \\
From the provided context (\textcolor{BurntOrange}{Document 1, Document 5, Document 7, Document 9}) we see that: \\
- Frank Sinatra, Paul Simon, and Stevie Wonder each have three wins. \\
- Taylor Swift has won the award four times: for *Fearless* (2010), *1989* (2016), *Folklore* (2021), and *Midnights* (2024). \\
Thus, \textcolor{BurntOrange}{Taylor Swift} holds the all‑time record for the most Album of the Year wins. \\
} \\

\midrule
\tabincell{p{0.95\linewidth}}{
\textbf{\ourGpt (low)}: \\
The answer can be found in the “Achievements” section of **\textcolor{BurntOrange}{Document 1} (Wikipedia – Grammy Award for Album of the Year)**. In that table the entry for *Taylor Swift* reads: Taylor Swift | 4 (2010, 2016, 2021, 2024) | Fearless (2008), 1989 (2014), Folklore (2020), Midnights (2022) | 7 (2010, 2014, 2016, 2021, 2022, 2024, 2025). The table also shows *Frank Sinatra* with 3 wins and other artists with up to 3, confirming that **no one has more than four wins**. \\
As a cross‑check, **BusinessInsider** article (\textcolor{BurntOrange}{Document 2}) states: "Taylor Swift… has won album of the year four times: Fearless (2010), 1989 (2016), Folklore (2021) and most recently Midnights (2024)." \textit{(54 words omitted for brevity)} ... The conclusion is that **\textcolor{BurntOrange}{Taylor Swift} holds the all‑time record for the most Album of the Year wins at the Grammys.**
} \\

\bottomrule
\end{tabular}
%%%%%%%%%%

    % }
    % \vspace{0.0em}
    \caption{\textbf{\dataset does not help models overcome reporting bias}. We show the response of \gptOssSmall and \ourGpt (with \texttt{low} reasoning effort) on another sample of LongSeal. Both models are inclined to {Taylor Swift} due to the \textcolor{BurntOrange}{reporting bias}, thus \textcolor{BurntOrange}{fail to reason} that the question does not target a specific role in music production.}
    \label{tab:not-good-qualitative-example}
    % \vspace{-1em}
\end{table}
% \end{wraptable}

To further understand the contribution of \dataset and its individual components, we conduct ablation studies on \gptOssSmall (low reasoning effort).

% \subsection{Reasoning style contributes little, while faithful reasoning patterns discovered by back translation are crucial for improvement.}
\subsection{Reasoning Style versus Faithful Reasoning Patterns}
We investigate whether forward sampling (with or w/o rejection) could serve as an effective alternative to backward translation for reasoning trace generation. Also, we re-evaluate the base model with a carefully-designed prompt (adapted from the prompt used for reasoning trace generation), to examine the contribution of the reasoning style encoded in \dataset.

Specifically, for forward sampling baselines, we prompt \gptOssLarge to generate reasoning traces for the 922 training samples. For the one w/o rejection, we directly finetune \gptOssSmall on these sampled reasoning traces regardless of their predicted answers' correctness. For the one with rejection, we compare the predicted answer against the ground-truth answer using the grader template of \dataset-Bench. Out of the 922 samples, 673 has reasoning traces passed the verification and retained for SFT. We finetune \gptOssSmall under the same configuration as our main experiment.

\begin{table*}[t]
    \small
    \centering
    \setlength{\tabcolsep}{4.5pt}     % narrow columns
    \renewcommand{\arraystretch}{1.1} % slightly taller rows
    % \resizebox{0.95\linewidth}{!}{%

%%%%%%%%%%
\begin{tabular}{lllllll}
\toprule
                        % & \multicolumn{3}{c}{\textit{In-domain}}           & \multicolumn{2}{c}{\textit{Out-of-domain}}                \\
                        % \cmidrule(lr){2-4}                                 \cmidrule(lr){5-6}
                        & LongSeal       & OfficeQA       & \dataset-Bench & LongBenchV2    & Oolong         & \textbf{Avg.}  \\

\midrule 
base                    & 34.38          & 22.38          & 60.86          & 38.76          & 27.96          & 36.87          \\
Back Translation (\dataset) & \textbf{39.69} & 30.40          & \textbf{68.01} & \textbf{44.52} & 32.98          & \textbf{43.12} \\

\midrule 
Sampling w/ Rejection   & 35.60          & 21.28          & 60.25          & 41.46          & 25.91          & 36.90          \\
Sampling w/o Rejection  & 35.83          & 19.42          & 63.43          & 43.66          & 29.37          & 38.34          \\
Prompting               & 35.04          & 20.00          & 63.48          & 42.79          & 26.57          & 37.58          \\

\midrule 
QA only                 & 18.55          & 14.29          & 54.49          & 36.28          & 27.46          & 30.21          \\
Table as context        & 34.25          & 20.61          & 62.36          & 37.67          & 23.74          & 35.73          \\
Expanded tables         & 35.43          & \textbf{30.53} & 65.71          & 44.34          & \textbf{37.17} & 42.64          \\

\midrule 
LongMIT                 & 28.46          & 22.22          & 55.93          & 37.85          & 22.55          & 33.40          \\

\bottomrule
\end{tabular}
%%%%%%%%%%

    % }
    % \vspace{0.0em}
    \caption{\textbf{Ablation study} on different SFT approaches with \gptOssSmall, and a comparison to a previous  long-context reasoning dataset LongMIT. The results show that \textit{faithful reasoning patterns grounded on realistic long context are crucial elements of \dataset}. Also, \dataset  closes the gap between previous long-context reasoning datasets and real-world demands.}
    \label{tab:ablation}
    % \vspace{-1em}
\end{table*}

As shown in \autoref{tab:ablation}, \textit{SFT on sampling, without or with rejection, yields substantially worse results than SFT with backward-translated traces}, as the average performance drops to 38.34/36.90 from 43.12. 
We attribute the performance degradations respectively to \textit{data correctness} and the \textit{selective nature of rejection sampling}. While there is no doubt that data correctness hinders the effective learning, discarding traces that produce incorrect answers also limit the learning to easier datapoints in the population. In contrast, \textit{backward translation} generates reasoning traces for all samples, including challenging ones, by conditioning on the given answer, thereby \textit{exposing the model to a broader spectrum of reasoning patterns}. This hypothesis is partially validated by the fact that, SFT on sampling w/o rejection perform slightly better than SFT on sampling with rejection, as the former is exposed to more challenging samples.

Meanwhile, there is no significant difference between the "Prompting" baseline and the original baseline. It suggests that the reasoning style does not have a significant contribution. In another words, faithful reasoning patterns learned from \dataset must be necessary factors for improvement.

% \subsection{Having more samples with expanded tables is not a must, but learning to reason with realistic context is crucial.}
\subsection{Contribution of Table Expansion and Context Realisticness}
We further ablate the table expansion, QA generation, and context components in \dataset to study their contributions. We compare \gptOssSmall and \ourGpt with the following baselines (with the same SFT setup to \ourGpt):

\begin{itemize}[leftmargin=*,label=$\bullet$,noitemsep,partopsep=0pt,topsep=0pt,parsep=0pt]
    \item \textit{Expanded tables}: As we notice that the proportion of samples with expanded tables in \dataset is modest, we generate more data and filter for samples with expanded tables only, until we obtain 1000 samples.
    \item \textit{QA only}: For each sample, we only use the question and the answer for finetuning. The long context and reasoning trace are not included.
    \item \textit{Table as context}: For each sample, we alter the free-form context by its table representation for reasoning trace generation and finetuning.
\end{itemize}

Results are shown in \autoref{tab:ablation}. The \textit{Expanded tables} baseline consistently performs better than the base model and is competitive to \ourGpt on some benchmarks. It suggests that though valuable, having more samples with expanded tables is not a must. We note that this result is not trivial, as samples with expanded tables are more challenging than samples with non-expanded tables and we expect \gptOssSmall can learn more from these samples than \dataset.

On the other hand, the \textit{QA only} baseline performs significantly worse than the base model across all benchmarks. It is surprising, as this finetuning means to give knowledge advantage over the base model while not affecting the model's reasoning behavior. Likewise, the \textit{Table as context} baseline slightly degrade from the base model. These results highlights that faithful reasoning grounded on realistic long context is the crucial element of \dataset.

% \subsection{\dataset closes the gap between previous long-context reasoning datasets and real-world demands.}
\subsection{Comparison to Previous Long-Context Reasoning Datasets}
We compare \dataset with an open-source reasoning dataset. We choose LongMIT~\citep{chen2025longmit} because of its accessibility and its focus on multi-hop question answering grounded in real documents. We finetune \gptOssSmall on the 800 training samples of LongMIT with the same configuration as our main experiment, and evaluate on the same benchmarks. As shown in \autoref{tab:ablation}, \textit{SFT on LongMIT yields worse results than both the base model and \ourGpt on all benchmarks}. This suggests that while LongMIT provides some benefits for long-context reasoning, it does not fully capture the complexity and diversity of reasoning patterns required for realistic applications, which \dataset is designed to address.

\section{Related Work} % DONE
\label{sec:relatedwork}

Despite rising demand for better long-context reasoning in LLM,
the availability of high-quality training data for such capabilities remains limited. Recent efforts address this gap through various approaches.

A line of works, including HotPotQA~\citep{yang2018hotpotqa}, MuSiQue~\citep{trivedi2022musique}, LongMIT~\citep{chen2025longmit}, CGMIS~\citep{sun2026cgmis}, and QwenLong~\citep{shen2025qwenlong}, focus on generating multi-hop question-answer pairs grounded in real documents. They used pre-defined reasoning patterns to structure the question answering samples from real documents, or parse the real documents into concept graphs, then generate question-answer pairs with verifiable reasoning traces.
However, the reasoning complexity of these datasets is limited by low branching factors of the reasoning trees, which restricts the diversity of reasoning paths and the breadth of inference required.

In contrast, IN2~\citep{an2024make}, LongFaith~\citep{yang2025longfaith}, and LoongRL~\citep{wang2025loongrl} attempted to generate long-context reasoning data by synthesizing long contexts where evidence is inserted at different positions. TableLong~\citep{xie2026probing} instead worked directly with tables, generating complex reasoning questions and training models directly in structured table data. 

Meanwhile, OpenR1~\citep{hf2025openr1} and OpenThoughts~\citep{guha2026openthoughts} provided high-quality reasoning data in Math, Coding, and Science by collecting questions from various sources, generating answers with LLMs, and filtering with meticulous schema. These approaches enable the generation of complex reasoning data with high branching factors, but the long contexts are mostly synthetic or limited to structured, easy-to-verify domains such as table question answering, natural science, and coding.

\textit{Our approach, \dataset, fills a missing piece by generating verifiable complex reasoning data with high branching factors, grounded in realistic long contexts in open-domain tasks, and guided by common data structures.} This allows us to systematically train LLMs to perform complex reasoning over realistic long contexts, improving their core reasoning capabilities.

\section{Conclusion}
\label{sec:conclusion}

We presented \dataset, a structure-based data curation pipeline that leverages Wikipedia tables to generate high-quality long-context reasoning data. Through dual-path verification and reasoning trace back-translation, our pipeline produces diverse complex questions with reliably correct answers over realistic long context. Supervised finetuning on \dataset yields consistent improvements, with \ourGpt and \ourQwen achieving +6.25\% and +3.37\% average gains respectively over the base models. Faithful reasoning patterns discovered by back translation and grounded realistic long context play a crucial role for the improvements. 
% By the direct comparison with a previous long-context reasoning dataset, we show that \dataset closes the gap between previous works and real-world demands. % We open source our pipeline, dataset, and models as fully open-source resources to facilitate reproducible research and future extensions. 

% DONE ALL below
% \clearpage
\section*{Ethics Statement}

We present a data curation method for improving reasoning capabilities in LLMs. We identify the following ethical considerations.

\textbf{Data source and privacy.} Our dataset is constructed from publicly available Wikipedia content and open-source LLMs. No private or personally identifiable information is included in our data pipeline or evaluation benchmarks. All used data and tools are openly accessible.

\textbf{Biases in synthetic data generation.} The reasoning traces and expanded table columns are synthetically generated by LLMs. While we implement verification steps to ensure factual correctness, synthetic data may inadvertently propagate biases present in the underlying model or source content. Wikipedia itself reflects historical and cultural biases that may be inherited in our dataset.

\section*{Acknowledgments}
As described in the main paper, we use LLMs as a part of our data pipeline and our benchmark's evaluation.
AI assistants (e.g., ChatGPT) are used to accelerate the coding, rephrasing, and grammar checking of our paper.

\clearpage
\bibliography{colm2026_conference}
\bibliographystyle{colm2026_conference}

\newpage
\appendix % DONE ALL below
% \note{Follow the presentation of s1 paper. https://arxiv.org/pdf/2501.19393}

\section{\dataset's Curation and Statistics}
\label{app:dataset}
% Topic distribution of original Wikipedia pages

\begin{table}[t]
    \small
    \centering
    \setlength{\tabcolsep}{6pt}     % narrow columns
    \renewcommand{\arraystretch}{1.1} % slightly taller rows
    % \resizebox{0.95\linewidth}{!}{%

%%%%%%%%%%
\begin{tabular}{ll}
\toprule
Topics                         & \% \\

\midrule
Reference and Event Pages      & 40.1       \\
Culture, Religion and Mind     & 16.4       \\
Science and Technology         & 12.4       \\
History, Politics and Conflict & 12.2       \\
Geography and Transport        & 8.8        \\
Entertainment and Media        & 4.5        \\
Law, Business and Institutions & 2.3        \\
Sports                         & 1.3        \\
Nature and Environment         & 1.1        \\
Education and Biography        & 0.8        \\

\midrule
Total                          & 100        \\

\bottomrule
\end{tabular}
%%%%%%%%%%

    % }
    % \vspace{0.0em}
    \caption{Distribution of Wikipedia pages' topics}
    \label{tab:wikipages-topic}
\end{table}

%%%%%%%%%%%%%%%%%%%%
% Quality score distribution

\begin{table}[t]
    \small
    \centering
    \setlength{\tabcolsep}{6pt}     % narrow columns
    \renewcommand{\arraystretch}{1.1} % slightly taller rows
    % \resizebox{0.95\linewidth}{!}{%

%%%%%%%%%%
\begin{tabular}{llll}
\toprule
LLM-judge Rating   & 3   & 4   & 5    \\

\midrule
Complexity         & 194 & 762 & 216  \\
Self-containedness & 0   & 123 & 1049 \\
Naturalness        & 0   & 354 & 838  \\

\bottomrule
\end{tabular}
%%%%%%%%%%

    % }
    % \vspace{0.0em}
    \caption{Distribution of LLM-judge automatic ratings of all 1,172 QA samples in \dataset. There are 68 training samples and 38 evaluation samples with all ratings of 5.}
    \label{tab:rating-distribution}
\end{table}

% \input{figures/data_statistics}
% \input{figures/topic_distribution_wikipedia_pages.tex}

% We show various data statistics in \autoref{tab:wikipages-topic}.

As stated, we choose Wikipedia as the source for table collection. That ensures high data quality, diversity, realisticness, and factual grounding.
Our dataset covers a variety of topics, with a large portion belonging to events (such as "2026 Winter Olympics"), as shown in \autoref{tab:wikipages-topic}. In the table collection step, we discard too small or too large tables (i.e., with fewer than 5 rows or 2 non-index columns, or more than 30 rows or 6 columns) to avoid simplicity or over-complication. In the step of QA generation in our data pipeline, a question is only kept if the automatic ratings for complexity, self-containedness, and naturalness are at least 3, 4, 4 respectively. Again, we elaborate the meaning of self-containedness as "providing all necessary parameters, definitions, and background information within itself, requiring no external links, data, references, or prior knowledge to be understood". \autoref{tab:rating-distribution} shows the distribution of automatic ratings of 1,172 validated QA samples in the whole \dataset dataset. For the automated web searching to form questions' contexts, we use LangSearch.\footnote{https://langsearch.com/}

To facilitate further understanding of our dataset, we show a sample reasoning trace generated by \qwenMedium for the question \texttt{\small Among the destinations where Uruguayan citizens can pick up pre-approved visas upon arrival, which country mandates that a sponsor must submit the visa application at least 48 hours prior to the traveler's arrival?} with the correct answer \texttt{\small Eritrea} and the main reference Wikipedia page \texttt{\small Visa requirements for Uruguayan citizens}.

\begin{rawprompt}[Sample Reasoning Trace]
* Step 1: Understanding the Question and Planning Strategy
The question asks to identify a specific country from a list of destinations where Uruguayan citizens can pick up "pre-approved visas upon arrival." The distinguishing criterion is that for this specific country, a **sponsor** must submit the visa application at least **48 hours prior** to the traveler's arrival. My strategy is to scan the provided documents for a section specifically discussing "Pre-approved visas" or "Visas pick-up on arrival," identify the list of countries mentioned there, and then check the specific conditions for each country to find the one matching the 48-hour sponsor requirement.

* Step 2: Locating Relevant Information in Documents
I need to find a document that lists countries allowing pre-approved visas to be picked up on arrival.
- Scanning **Document 1** (Live in Uruguay): This document details entry procedures, visas, and residency types for entering Uruguay, but it does not list countries where *Uruguayans* can pick up visas. It focuses on incoming travel to Uruguay.
- Scanning **Document 3** (ETIAS for Uruguayan Citizens): This discusses ETIAS requirements for Europe, not general pre-approved visas for other countries.
- Scanning **Document 4** (Wikipedia - Visa requirements for Uruguayan citizens): I see a section titled "Pre-approved visas pick-up". Let's examine this section closely.
- Scanning **Document 5** (Visa Requirements for Uruguayan Citizens): This document categorizes countries by visa type (Not Required, On Arrival, eVisa) but does not have a specific "Pre-approved visas pick-up" section with detailed conditions like the Wikipedia page.
- Scanning **Document 6** (Administrative entry restrictions): This appears to be a duplicate or very similar version of Document 4. It also contains a "Pre-approved visas pick-up" section.

Let's focus on **Document 4** (and confirmed by **Document 6**) as they explicitly contain the section header "Pre-approved visas pick-up".

* Step 3: Extracting Candidates from the "Pre-approved visas pick-up" Section
In **Document 4**, under the section "*Section*: Pre-approved visas pick-up", the text states: "Pre-approved visas can be picked up on arrival in the following countries instead in embassy or consulate."
The table lists the following countries:
1.  Bhutan
2.  Cameroon
3.  Eritrea
4.  Liberia
5.  Nigeria
6.  Sudan
7.  Turkmenistan

Now I must check the specific condition for each of these candidates to see which one requires a sponsor to apply 48 hours in advance.

* Step 4: Evaluating Candidates Against the Criteria
Let's examine the conditions listed for each country in the table found in **Document 4** (and verified in **Document 6**):

-   **Bhutan**: The condition is "For a maximum stay of 15 days if the application was submitted at least 2 and a half months before arrival and if the clearance was obtained."
    -   *Analysis*: This mentions a time frame (2.5 months), but it refers to the applicant submitting the application, not necessarily a sponsor, and the timeframe is different (months vs hours). This does not match the "48 hours prior" criterion.

-   **Cameroon**: The condition is "Must hold approval from the General Delegate of Security."
    -   *Analysis*: No mention of a sponsor or a 48-hour deadline.

-   **Eritrea**: The condition is "Must have a sponsor who must submit an application at least 48 hours before arrival."
    -   *Analysis*: This matches all criteria perfectly.
        -   Requires a **sponsor**.
        -   The sponsor must **submit the application**.
        -   The deadline is **at least 48 hours before arrival**.

-   **Liberia**: The condition is "Available only if arriving from a country without a diplomatic mission of Liberia and if a sponsor obtained an approval."
    -   *Analysis*: Mentions a sponsor, but no specific 48-hour submission deadline is listed.

-   **Nigeria**: The condition is "Holders of a visa application who have a Nigerian company taking responsibility for them."
    -   *Analysis*: Mentions a company (similar to a sponsor), but no specific 48-hour timeline is mentioned.

-   **Sudan**: The condition is "Holders of an entry permit issued by the Ministry of Interior."
    -   *Analysis*: No mention of a sponsor or 48-hour rule.

-   **Turkmenistan**: The condition is "Holders of an invitation letter of the local company that was approved by the Ministry of Foreign Affairs."
    -   *Analysis*: Mentions a company/invitation, but no 48-hour deadline.

* Step 5: Cross-Verification
Let's verify this finding using **Document 6** ("Administrative entry restrictions"), which contains the same table.
Under the section "*Section*: Pre-approved visas pick-up":
| Country | Conditions |
| :--- | :--- |
| ![](image) Eritrea[326] | Must have a sponsor who must submit an application at least 48 hours before arrival. |

The information is consistent across both documents. The specific phrasing "Must have a sponsor who must submit an application at least 48 hours before arrival" is unique to Eritrea in this list.

* Step 6: Final Conclusion
Based on the analysis of the "Pre-approved visas pick-up" section in Document 4 and Document 6, the only country that mandates a sponsor to submit the application at least 48 hours prior to arrival is Eritrea.

* Step 7: Answer Formulation
The correct answer is Eritrea.
\end{rawprompt}

\section{Training Details}
\label{app:training}

We finetune two base models on our \dataset training data: \gptOssSmall (a 20B parameter model using MXFP4 quantization) and \qwenSmallInst (a 4B parameter instruction-tuned model). Both models are loaded via Unsloth for memory efficiency.

\begin{figure*}[t]
\centering
\resizebox{\linewidth}{!}{%
    \includegraphics[ width=\textwidth]{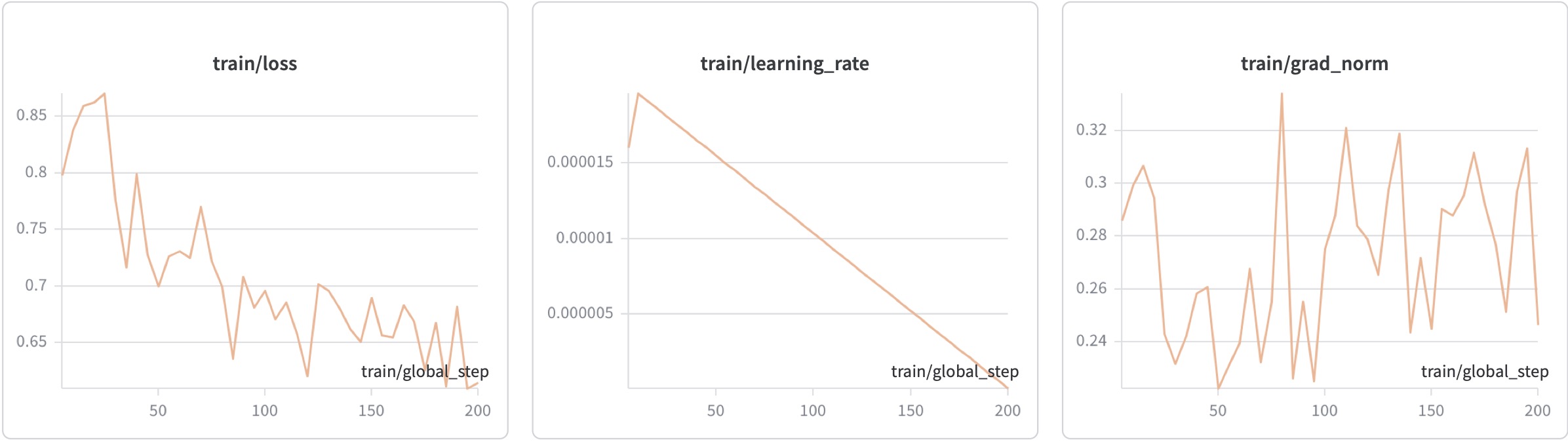} 
}
\caption{Training dynamics of \gptOssSmall on \dataset.}
\label{fig:training-dynamics}
% \vspace{-1em}
\end{figure*}

We use LoRA~\citep{hu2022lora} with rank $r=8$ and target modules $\{q, k, v, o\}$ for attention layers, with alpha set to $2r$. Training is performed with a total batch size of 4 and a learning rate of $2e{-5}$ with linear warmup for 5 steps followed by linear decay. The batch size per device and the gradient accumulation depend on the availability of hardware. We train for 200 steps, approximately 1 epoch over our 922 training samples. The maximum sequence length is set to 128K tokens to accommodate our long reasoning traces. We use the AdamW optimizer with weight decay of 0.001 and train in mixed precision (bfloat16 for \qwenSmallInst, 4-bit quantization for \gptOssSmall). The training takes approximately 2.5 hours on 4 NVIDIA H200 GPUs using PyTorch DDP. \autoref{fig:training-dynamics} shows the training dynamics of \gptOssSmall on our dataset.

We format training samples using chat templates with special token delimiters. For \gptOssSmall models, we use reasoning effort tokens (\texttt{low} or \texttt{high}) to control the thinking budget; for \qwenSmallInst models, we use the standard \verb@<|im_start|>@ format.
% The system prompt instructs models to extract and aggregate necessary information from the provided context before reasoning step by step to arrive at the final answer.
We compute loss only on the assistant response (reasoning traces and final answer), not on the user prompt or system message.

Note that for \gptOssSmall, evaluation uses the same reasoning effort configuration as training. That is, \ourGpt with \texttt{high} reasoning effort was trained with \texttt{high} reasoning effort and evaluated with \texttt{high} reasoning effort; similarly for \texttt{low}. This ensures consistency between the learned reasoning patterns and the evaluation setup.

\section{Additional Experiments and Analyses}
\label{app:experiment}
% Ablation with 100 best quality samples (excluding the held-out evaluation set) and 
% Ablation with self-distillation

\begin{table*}[t]
    \small
    \centering
    \setlength{\tabcolsep}{4.5pt}     % narrow columns
    \renewcommand{\arraystretch}{1.1} % slightly taller rows
    % \resizebox{0.95\linewidth}{!}{%

%%%%%%%%%%
\begin{tabular}{lllllll}
\toprule
                    & LongSeal       & OfficeQA       & \dataset-Bench & LongBenchV2    & Oolong         & \textbf{Avg.}  \\

\midrule
\multicolumn{6}{l}{\gptOssSmall \textit{Low reasoning effort}} \\
base                & 34.65          & 21.63          & 62.92          & 39.53          & 31.03          & 37.95          \\
\ourGpt             & 38.19          & 26.53          & \textbf{67.05} & 46.05          & \textbf{33.31} & 42.23          \\
SFT on \dataset-100 & \textbf{40.94} & 18.75          & 63.48          & 47.57          & 25.72          & 39.29          \\
Self-distillation   & 37.8           & \textbf{28.89} & 66.29          & \textbf{48.45} & 30.34          & \textbf{42.35} \\
% SFT with RS         & 35.6           & 21.28          & 60.25          & 41.46          & 25.91          & 36.9           \\

\midrule
\multicolumn{6}{l}{\gptOssSmall \textit{High reasoning effort}} \\
base                & 52.17          & 37.84          & 72.57          & 46.07          & 35.86          & 48.9           \\
\ourGpt             & \textbf{64}    & \textbf{46.58} & \textbf{77.27} & 46.32          & 31.95          & \textbf{53.22} \\
SFT on \dataset-100 & 58.01          & 41.94          & 74.21          & \textbf{51.7}  & 32.9           & 51.75          \\
Self-distillation   & 55.56          & 29.33          & 72.12          & 44.38          & 35.9           & 47.46          \\
% SFT with RS         & 44.09          & 23.08          & 63.48          & 39.77          & \textbf{41.71} & 42.43          \\

\bottomrule
\end{tabular}
%%%%%%%%%%

    % }
    % \vspace{0.0em}
    \caption{\textbf{Ablation study} with \gptOssSmall (both \texttt{low} and \texttt{high} reasoning efforts) on a minimal \dataset-100 training set and self-distillation. %, and SFT with rejection sampling (RS).
    \textbf{Bold-faced numbers} indicates the best among the group. \dataset-100 brings improvements to both reasoning effort configuration, though the improvements are less than those from the full \dataset training set. Meanwhile, self-distillation shares similar improvements with our original setup for \texttt{low} reasoning effort configuration, but does not hold for \texttt{high} reasoning effort configuration. % Notably, we found that SFT with RS on \dataset is not effective as our default SFT with back translation.
    }
    \label{tab:ablation-appendix}
    % \vspace{-1em}
\end{table*}

\begin{table*}[t]
    \small
    \centering
    \setlength{\tabcolsep}{4.5pt}     % narrow columns
    \renewcommand{\arraystretch}{1.1} % slightly taller rows
    % \resizebox{0.95\linewidth}{!}{%

%%%%%%%%%%
\begin{tabular}{lllllll}
\toprule
                    & LongSeal       & OfficeQA       & \dataset-Bench & LongBenchV2    & Oolong         & \textbf{Avg.}  \\

\midrule 
\gptBest            & n/a            & n/a            & 81.46          & n/a            & n/a            & n/a            \\
% \gptMini \\
% \gptOldMini \\

\midrule \midrule
\multicolumn{6}{c}{\qwenSmallInst} \\
\midrule
base                & 33.07          & 14.88          & 56.18          & 37.56          & 29.67          & 34.27          \\
\textbf{SFT, lr=2e-5 (ours)}
                    & 40.94          & 13.58          & 55.62          & 39.25          & 35.48          & 36.97          \\
% \; infer. w/ $T=0$  & 36.61          & 16.33          & 54.49          & 40.19          & 35.01          & 36.53          \\
\midrule
SFT, lr=2e-6        & 29.13          & 13.47          & n/a            & 37.85          & 30.82          & n/a            \\
SFT, lr=2e-4        & 28.93          & 53.09          & n/a            & 30.52          & 29.6           & n/a            \\
\midrule
SFT noisy context   & 36.22          & 16.73          & 54.49          & 36.45          & 33.28          & 35.43          \\
\midrule
RLM*                & 28.85          & 28.57          & 53.71          & 28.57          & 48.95          & 37.73          \\
SFT + RLM*          & 34.54          & 21.43          & 49.38          & 32.54          & 40.51          & 35.68          \\
\midrule
Prompting           & 30.71          & 15.98          & 54.49          & 32.82          & 29.91          & 33.88          \\

\midrule \midrule
\multicolumn{6}{c}{\gptOssSmall \textit{Low reasoning effort}} \\
\midrule
base                & 34.65          & 21.63          & 62.92          & 39.53          & 31.03          & 37.95          \\
\textbf{SFT, lr=2e-5 (ours)} 
                    & 38.19          & 26.53          & 67.05          & 46.05          & 33.31          & 42.23          \\
% \; infer. w/ $T=0$  & 39.76          & 28.63          & 65.12          & 46.53          & 31.74          & 42.36          \\
\midrule
SFT, lr=2e-6        & 40.16          & 22.04          & n/a            & 39.07          & 28.22          & n/a            \\
SFT, lr=2e-4        & 54.97          & n/a            & n/a            & 38.78          & n/a            & n/a            \\
\midrule
SFT noisy context   & 40.55          & 24.18          & 65.73          & 44.19          & 36.11          & 42.15          \\
\midrule
RLM*                & 45.28          & 31.82          & 63.48          & 34.04          & 41.55          & 43.23          \\
SFT + RLM*          & 49.21          & 31.82          & 64.2           & 29.29          & 38.55          & 42.61          \\
\midrule
Prompting           & 35.04          & 20.00          & 63.48          & 42.79          & 26.57          & 37.58          \\

\bottomrule
\end{tabular}
%%%%%%%%%%

    % }
    % \vspace{0.0em}
    \caption{\textbf{Additional experiments}. Due to the prohibitively expensive cost of GPT-5.4, we only run experiment on our \dataset-Bench. \textbf{Rows respectively represent}: 
    (1) \textit{base}: the base model, 
    (2) \textit{SFT, lr=2e-5 (ours)}: our finetuned model with default setup, 
    % (3) \textit{infer. w/ $T=0$}: our finetuned model with default setup except that the inference temperature is 0, 
    (3) \textit{SFT, lr=2e-6}: our finetuned model with a smaller learning rate 2e-6,
    (4) \textit{SFT, lr=2e-4}: our finetuned model with a larger learning rate 2e-4,
    (5) \textit{SFT noisy context}: our finetuned model with data which the context is naively prepared,
    (6) \textit{RLM}: Recursive Language Models,
    (7) \textit{SFT + RLM}: the combination of our SFT and RLM, where we first apply RLM on top of our finetuned model for inference, and 
    (8) \textit{Prompting}: the base model but evaluated with a dedicated prompt adapted from the reasoning trace generation prompt.
    \textbf{n/a} entries indicate that the model's performance could not be reliably measured due to excessive inference time, or unconsidered evaluation after hyperparameters probing.
    \textit{The results suggest that our SFT with the default setup (lr=2e-5) still yields the best performance}, while other variants of SFT and RLM show mixed results.
    (*) We note that the result of Recursive Language Models (RLM) is more than 50\% incomplete due to the context overflow problems of the \texttt{rlms==0.1.1} package that we encountered. Thus, although its results are competitive, the comparison between its and ours is not apple-to-apple.}
    \label{tab:exp-appendix}
    % \vspace{-1em}
\end{table*}

% \gptOssSmall &  & \qwenSmallInst &  & Others &  \\
% \multicolumn{2}{c|}{\gptOssSmall} & \multicolumn{2}{c|}{\qwenSmallInst} & \multicolumn{2}{c}{Others} \\

% [DO THE BELOW LATER]
% Train on cross-model-family reasoning trace
% Reasoning trace: gpt-oss-120b vs. Qwen3.5-35B-A3B generated?

Beyond the main experimental results presented in \S\ref{sec:experiments}, we conducted several additional studies to better understand the properties of our finetuned models and the effectiveness of \dataset. These supplementary experiments are summarized in \autoref{tab:ablation-appendix} and \autoref{tab:exp-appendix}.

\subsection{Efficiency of High-Quality Data.}
We investigate whether a small set of our highest-quality samples can boost model performance. Specifically, we finetune \gptOssSmall 25 steps on just 100 samples with the sum of quality ratings at least 14 over 15, selected from the \dataset's training set. As shown in \autoref{tab:ablation-appendix}, even with this minimal subset (\dataset-100), the finetuned model achieves noticeable improvements over the base one: +1.34 points average with \texttt{low} reasoning effort (39.29 vs. 37.95) and +2.85 points with \texttt{high} reasoning effort (51.75 vs. 48.90). While these gains are smaller than those achieved with the full 922-sample training set, they demonstrate that our curation pipeline produces high-quality data where even a small subset provides meaningful supervision. This validates the efficiency of \dataset, as reasonable improvements can be achieved with minimal computational cost.

\subsection{Effective Self-Distillation with Curated QA Data.}
A feature of \dataset is that the reasoning traces are generated by large teacher models (\gptOssLarge and \qwenMedium). Here, we investigate whether smaller models can benefit from their own generated traces through self-distillation. We use the base \gptOssSmall to generate reasoning traces for QA samples and finetune it on these self-generated traces. As shown in \autoref{tab:ablation-appendix}, self-distillation yields impressive results with \texttt{low} reasoning effort: the average performance improves from 37.95 (base) to 42.35 (+4.40), competitive with our original SFT setup (42.23). This demonstrates \textit{the value of our core QA curation} that facilitates effective self-distillation. However, with \texttt{high} reasoning effort, self-distillation shows no improvement performance (47.46 vs. 48.9). It indicates that the thinking patterns scaffolded by \gptOssSmall itself are not new to the model with \texttt{high} reasoning effort, unlike those scaffoled by the larger \gptOssLarge. Our results suggest the importance of using stronger teacher models for effective learning in reasoning-intensive models.

\subsection{Effect of Learning Rate}

We investigate the impact of learning rate (lr) on model performance during supervised finetuning. In addition to our default learning rate of $2e{-5}$, we experiment with both smaller ($2e{-6}$) and larger ($2e{-4}$) learning rates. As shown in \autoref{tab:exp-appendix}, the smaller learning rate of $2e{-6}$ possibly results in underfitting, where the model fails to learn sufficient reasoning patterns from \dataset and only performs comparably to or marginally better than the base model. 
% Specifically, for \gptOssSmall with \texttt{low} reasoning effort, the average performance in four considered benchmarks with lr=$2e{-6}$ reaches only 32.37, marginally above the base of 31.69. Similarly, \qwenSmallInst with lr=$2e{-6}$ achieves an average of 27.82, which is a small fraction better than 27.48 of the base model, while being substantially lower than \ourQwen.

Conversely, the larger learning rate of $2e{-4}$ causes training instability, even when setting the weight decay of 0.01 and 0.1. We hypothesize that this case is akin to the \textit{alignment reversal} phenomenon described in \citet{jxm2025ossbase}, where the finetuned model's weights diverge substantially from the base model, leading to unexpected generation behaviors. In particular, the model produces excessively long outputs that frequently cause API timeout errors during evaluation, making reliable performance measurement impossible. These findings underscore the importance of careful learning rate selection for long-context reasoning finetuning, and we chose lr=2e{-5} as our default after initial hyperparameter probing.

\subsection{Effect of Context Cleanliness}

As mentioned in \S\ref{sec:data}, we implement a strict validity check to ensure that all evidence remains present within the final context for reasoning trace generation. Here, we study possible benefit of the reversely defined noisy context setup, where we do not explicitly keep evidence passages during the compaction. That mimics noisy real-world QA problem, as some explicit evidences to be missing in the final context, and the model must resort to alternative sources of information scattered across the context or embedded in model weights. The results, labeled as "SFT noisy context" in \autoref{tab:exp-appendix}, suggests that using the clean context setup is still preferred, though the observed gaps are not statistically significant.

\subsection{Validation of \dataset-Bench}

Our held-out benchmark \dataset-Bench serves both as an evaluation tool and a validation of our data curation pipeline's quality. The \gptBest performance confirms that the curated questions are solvable and still having room for improvement. After finetuning base models on \dataset, performance on \dataset-Bench improves from 62.92 to 67.05 for \gptOssSmall, demonstrating that the reasoning patterns learned from \dataset transfer effectively to our curated benchmark.

\subsection{Comparison with Recursive Language Models}

Recursive Language Models (RLM)~\citep{zhang2025recursive} represent an orthogonal approach to handling long-context reasoning through recursive context processing. We compare our SFT approach with RLM and investigate potential synergies between the two methods. For RLM experiments, we use the publicly available \texttt{rlms==0.1.1} package and apply it to both base models and our finetuned models during inference. We note that the result of RLM is more than 50\% incomplete, due to a context overflow problem of the current \texttt{rlms==0.1.1} package that we encountered. Thus, the comparison between RLM and our models are not apple-to-apple.

As shown in \autoref{tab:exp-appendix}, RLM applied to base \gptOssSmall achieves an average of 43.23, which is competitive with our SFT approach (42.23). Notably, RLM shows particular strength on Oolong (41.55 vs. 33.31 for SFT), suggesting that recursive processing is effective for distributional reasoning tasks. When we apply RLM on top of our SFT models ("SFT + RLM"), the results are mixed, as some benchmarks show improvement while others degrade. That is explainable, as our finetuned model may have different instructional behaviors from the base one that affects its use of RLM pipeline. Overall, our findings suggest that RLM and SFT on \dataset address long-context reasoning through different mechanisms, and a naive combination does not yield consistent synergistic benefits.

\begin{figure*}[t]
\centering
\resizebox{0.9\linewidth}{!}{%
    \includegraphics[draft=false, width=\textwidth]{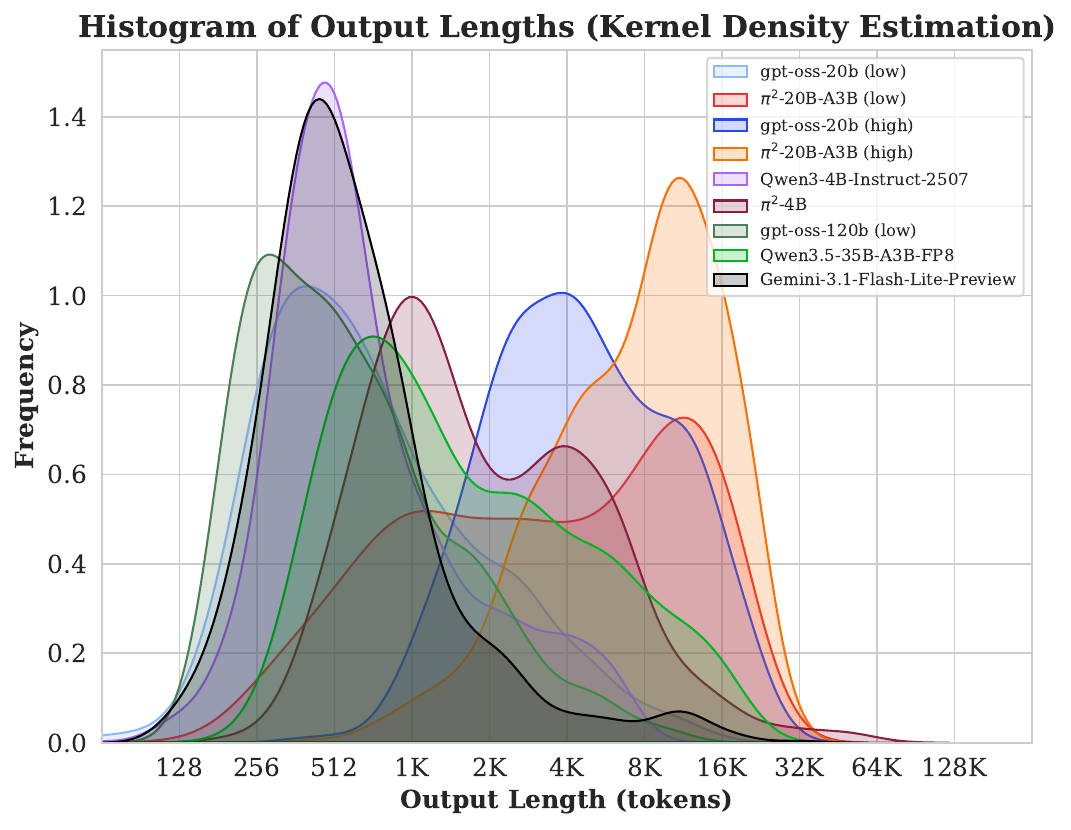}
}
\caption{Comparing output length (in tokens) over all benchmarks of our models, base models, larger open-source models, and leading commercial \geminiFlashLite. Here, the token count includes thinking tokens. Our models generate relatively more tokens than the base models (so-called \textit{overthinking tax}). Interestingly, our \ourQwen shares a similar output length distribution as its teacher model \qwenMedium, suggesting successful reasoning skill transfer. Meanwhile, \geminiFlashLite desirably generate the least tokens among considered models and have the best performance. That is what we ultimately aim for.}
\label{fig:output-length}
% \vspace{-1em}
\end{figure*}

\subsection{Output Length Analysis}
\label{app:length-analysis}

Given that our training data contains relatively long reasoning traces (median 1,480 tokens), we analyze whether finetuning induces changes in output behavior. \autoref{fig:output-length} compares the output token distributions (including thinking tokens) across base models, our finetuned models, larger open-source models, and commercial baselines.

The analysis reveals that our finetuned models consistently generate longer outputs than their base counterparts. For \gptOssSmall, the output length increases substantially after SFT, especially for \texttt{low} reasoning effort. On the other hand, \ourQwen exhibits an output length distribution remarkably similar to its teacher model \qwenMedium, suggesting that our data pipeline successfully transfers reasoning style from larger to smaller models. Meanwhile, \geminiFlashLite desirably generate the least tokens among considered models and have the best performance. That is what we ultimately aim for.

While longer outputs are not inherently desirable, they reflect the complex reasoning patterns learned from our training data, as shown in \autoref{tab:good-qualitative-example}. We note that in some samples, fine-tuned models exhibit overthinking or over-verbose, which represents an area for future improvement in balancing thoroughness and conciseness.

% \subsection{Evaluation of Zero-shot Baselines with a Dedicated Prompt}

% By default, for each benchmark, we use its default prompt setup. As the improvement brought by SFT with back-translated reasoning traces can be attributed to both new reasoning patterns (e.g., logic) and new reasoning style (e.g., carefulness), we ask: \textit{How much does the reasoning style encoded in the back translation prompt contribute to the improvement, apart from new reasoning patterns in our data?}

% We thus evaluate the base models with a dedicated prompt that is similar to the one used for generating reasoning traces in our data pipeline. As shown in \autoref{tab:exp-appendix}, the results with the dedicated prompt are largely consistent with the default evaluation settings, with minor variations. That suggests that the improvement brought by SFT with back-translated reasoning traces is not because of the reasoning style encoded in the prompt, but mainly the new reasoning patterns learned from our training data.

\subsection{Statistical Significance of Results}

% Evaluation Determinism (ref: s1 paper) and
% To ensure reproducibility and assess the reliability of our findings, we conduct experiments under deterministic evaluation settings. Specifically, we evaluate our finetuned models with temperature $T=0$, which eliminates random sampling effects. As shown in \autoref{tab:exp-appendix}, the results with $T=0$ are largely consistent with our default evaluation settings, with minor variations.

Following the approach of \cite{muennighoff2025s1}, we conduct a statistical significance t-test using 10,000 bootstrap samples drawn from all benchmark results. The improvements achieved by our finetuned models over base models are statistically significant at $\alpha = 0.01$ for both \gptOssSmall (under both reasoning effort configurations) and \qwenSmallInst, providing strong evidence that the observed performance gains are genuine and not attributable to random variation.

\section{Attempts at Larger Improvements}
\label{app:further-attempts}

\begin{table}[t]
    \small
    \centering
    \setlength{\tabcolsep}{4.5pt}     % narrow columns
    \renewcommand{\arraystretch}{1.1} % slightly taller rows
    % \resizebox{0.95\linewidth}{!}{%

%%%%%%%%%%
\begin{tabular}{lllll}
\toprule
FT & LR & Weight Decay & Steps & Avg. \\

\midrule
\multicolumn{5}{l}{\textit{Probing with \gptOssSmall High reasoning effort}} \\
SFT & 2e-4 & 0.1  & 200 & n/a \\
SFT & 2e-5 & 0.01 & 200 & n/a \\
SFT & 5e-6 & 0.01 & 200 & 53.5 \\

\midrule
\multicolumn{5}{l}{\textit{Scaling with \gptOssSmall High reasoning effort}} \\
SFT & 5e-6 & 0.01 & 500 & n/a \\
SFT & 5e-6 & 0.01 & 1000 & n/a \\

\midrule
\multicolumn{5}{l}{\textit{Trials with \qwenSmallInst}} \\
SFT & 2e-4 & 0.1  & 500 & 30.72 \\
SFT & 2e-5 & 0.01 & 500 & 36.71 \\
GRPO & 2e-5 & 0.01 & 500 & 31.68 \\

\bottomrule
\end{tabular}
%%%%%%%%%%

    \caption{\textbf{Attempts at Larger Improvements}. Additional experiments probing learning rate, scaling up training steps with \gptOssSmall, and trials with \qwenSmallInst. Column "FT", "LR", "Avg." respectively denotes the finetuning method, learning rate, and average performance of each experiment. The \textit{n/a} entries indicate that the model's performance could not be reliably measured due to excessive inference time. Overall, our attempts in scaling up the training time with more data did not yield further improvements.}
    \label{tab:further-attempts}
\end{table}

% Scaling up the data size and the training time
Because our data pipeline is fully open source and thus our dataset can be scaled up naturally, following ~\cite{guha2026openthoughts}, we attempted at larger improvements through scaling up the training data.
We have further processed 30000+ pages from Wikipedia, having totally 80000+ qualified tables and 14000+ validated QA samples with reasoning traces. In this section, we mostly work with \gptOssSmall (\texttt{high} reasoning effort), as it is relatively the best base model we finetuned in this paper.

\textbf{Probing hyperparameters}: We observed significant increase in completion tokens count after finetuning, even excessively so that the generation are not able to complete (as SFT with learning rate $2e{-4}$). We attributed the problem to weight divergence of the finetuned model and the base model. Thus, we first optimized learning rate and weight decay to avoid that problem in longer finetuning. We fixed the optimizer as AdamW with gradient clipping at 1.0, and the learning rate scheduler as linear warmup for 5 steps followed by linear decay. We also fixed the total number of finetuning steps to 1500, stopping step at 200, global batch size at 4, and LoRA rank at 8. Results are shown in \autoref{tab:further-attempts}. Among probed learning rate and weight decay pairs, we find that only the learning rate of 5e-6 and weight decay of 0.01 lead to an overall improvement which is aligned with our main experiments (53.5 vs. 53.22) while having the inference time reasonable. Thus, we use that pair for scaling up the training.

\textbf{Scaling up training steps}: However, further increasing the training steps to 500 or 1000 with the chosen setup again leads to excessive inference time. That is likely because the model weights diverge more from the base model as training goes on, which leads to more verbose generation.

\textbf{Trials with \qwenSmallInst}: We also attempted at improving \qwenSmallInst with more training steps and with GRPO. For GRPO, we adapt \dataset-Bench's LLM Grader prompt to a rule-based function to compute reward of rollouts. While we did not encounter excessive inference time problem with \qwenSmallInst, the results are not better than the default SFT setup in our main experiments. For the GRPO-finetuned model, we observe a stable training reward from the start to the end of the training process.

\textbf{Possible explanation}: While we will further investigate the excessive inference time problem of \gptOssSmall in future work, we would like to comment on the result of \qwenSmallInst. The above results suggest that simply scaling up the training time with more data does not necessarily lead to larger improvements. We hypothesize that the difficulty and task diversity of the training data are also key factors in driving performance gains. Our dataset is limited in the question type (by design) and limited in difficulty (as the evaluation on \dataset-Bench showed). Thus, we believe that our future work should focus on improving the difficulty and diversity of training samples.

% \textbf{Model finetuning.} Our finetuning setups are limited to relatively small-size language models, SFT with LoRA paradigms, and with a limited set of hyperparameters. More extensive exploration of training strategies (e.g., PPO, GRPO) and hyperparameter tuning may yield further improvements.

\section{Limitations}
% \note{COLM does not require Limitations section, not sure if they count towards page limit. We may not use page beyond the 9th page to add one. Consider move to Appendix}

While our work demonstrates strong empirical results, several limitations should be acknowledged.

\textbf{Context length scope.} Our study focuses on long-context reasoning within the 128K-token context range, which covers common context window limits but does not address extremely long contexts (e.g., million-token scale). While our pipeline can theoretically generate data for longer contexts, we leave exploration of this regime for future work.

\textbf{Quality and yield.} Despite our dual-path verification and quality filtering, approximately 17\% of initially generated samples required human revision in our benchmark curation. While this demonstrates the effectiveness of our pipeline, it also indicates room for improvement in automatic quality control.

\textbf{Evaluation setup.} We evaluate models on long-context reasoning tasks in context-provided QA setup. Performance on other long-context capabilities (e.g., retrieval accuracy, long-context free-form generation) is not directly assessed. % Also, more statistics about our \dataset-Bench will give readers a clearer picture of this new benchmark.

\section{Prompts}
\label{app:prompt}
We show the prompts used in our experiments in this section.

\noindent\rule{\linewidth}{0.5pt}

\textbf{Table expansion}

\begin{rawprompt}[]
You are a creative and grounded brainstorming partner for enriching tables in Wikipedia pages.

You will be given:
- a table in HTML format,
- some metadata about the table,
- summaries of the Wikipedia pages referenced by a column of the table.

Your task is to enrich the table by adding one or several (up to 3) new columns that capture common attributes from the Wikipedia summaries which are not already semantically reflected in the existing columns of the table.

**Action plan**:
- Identify one or several common attributes from the Wikipedia summaries. Each attribute must be:
    - relevant to the data in the table and provide additional insights to the readers,
    - **not already present or partially reflected in the existing columns of the table**,
    - not in conflict with the existing data in the table (e.g., statistics from different years compared to similar existing attributes).
- Extract the relevant data/values from the summaries to populate the new columns.
- Values must be short pieces of content (e.g., a number, a phrase, a several-word string), directly retrieved from the summaries, no extra notes, and should not require additional interpretation or analysis.
- Return the enriched table in HTML format, ensuring that the new columns are seamlessly integrated with the existing columns.
- **Provide a list of notes on the enriched table** about any cautions or assumptions made during the enrichment process, for examples, in the follow cases:
    - Statistics in a new column are from different years, so they may not be directly comparable. You should provide such a note with specific year information, e.g., "- Attribute X: row 1's value is from 2020, while row 2's value is from 2022."
- In case there are typos in the original table, you are allowed to modify these typos. Don't need to report the incident. Typos cases include but are not limited to:
    - Misspelled words -> Fix the typos (e.g., "Spanyol" -> "Espanyol")
    - Inconsistent (and not sensical) statistics -> Resolve the inconsistent or the typos (e.g., in a numeric column, values are in different units, or one uses `,` as thousand separator while others use `.`, etc.)

NOTE:
- Existing column names may not fully reflect the content of the column, so you should also consider the actual data in the columns when determining whether an attribute is already present.
- There are some common taxonomies of attributes that are often present in Wikipedia summaries, such as:
    - Person: teams/companies/organizations, awards, career milestones, etc.
    - Organization: industry, headquarters location, revenue, number of employees, etc.
    - Event: date, location, participants, outcome, etc.
    - Geographical entity: population, area, GDP, neighboring entities, etc.
    - Knowledge: year, field, key contributions, inventors, etc.

Now, enrich the table based on the above instructions and the following input.

[[ ## table ## ]]
{table}

[[ ## metadata ## ]]
{metadata}

[[ ## column_name ## ]]
{column_name}

[[ ## summaries ## ]]
{summaries}
\end{rawprompt}
\noindent\rule{\linewidth}{0.5pt}

\textbf{QA generation}

\begin{rawprompt}[]
You are an expert designed for advanced data analysis.

You are given a table in Markdown format, a column name, and some context about the table, which reflect real information from the Internet.
Your task is to generate **a complex, self-contained, and natural question** about the data along with a SQL implementation to answer it.

**Strict Requirements:**

1. **Question Design:**

1.1. Generate ONE single-focus and concise question.
    - DON'T concatenate multiple sub-questions with "and" or "which... and which".
    - Split multi-part questions into separate questions. For example, instead of "which X and which Y", ask only "which X". 
    - Instead of "which X and which Y when X", just ask "When X, then which Y?" to maintain the complexity and multi-hop reasoning, but avoid asking two separate questions in one.
        - DON'T: "Which team has the highest average score and which team has the lowest average score?", "What is the year when the Lansing fare per mile reached its highest recorded value, and what was the difference in fare per mile between Lansing and Grand Rapids in the year?"  
        - DO: "Which team has the highest average score?", "Which team has the lowest average score?", "In the year when the Lansing fare per mile reached its highest recorded value, what was the difference in fare per mile between Lansing and Grand Rapids?"

1.2. **Complexity**: Ensure the question probes deeply into the data's nuances.
    - It should involve multi-hop reasoning and information aggregation, by embedding multiple constraints or conditions in the question.
        - Example: "in the year when X reached its maximum", "for the team with the lowest average score in X", "among the teams with average score above 80", etc.
    - It can involving (conditional) **counting or information aggregation (e.g., sum, max, min, average, etc.)** to increase the complexity.
    - Nonetheless, the question is still as meaningful and natural as being asked by real data analysts.

1.3. **Self-containedness**: The question should be fully self-contained, providing all necessary context and details within the question itself, without requiring external information or references to the table or metadata.
    - Utilize specific information in the table and metadata to design the question. Assuming that the question will be given without any context, and one can still fully understand it.
        - For example, given a table with a section name "Number of athletes by National Olympic Committee". phrase "ranked 31 to 45" should be more specific to explain which ranking is referred to, such as "ranked 31 to 45 in the number of athletes sent to the Olympics".
    - Must have phrases about time, location, or taxonomy etc. to narrow down the scope of the question.
        - Example: "In the year when X reached its highest value", "In competition X, among the teams with average score above 80", "for the team with the lowest average score in X", etc.
    - The wording **should NOT refer to the table or any metadata items**, as real users do NOT have access to these information when being asked.
        - AVOID phrases like "which column", "in the column of", "the value in column X", or, "as listed in (somewhere), "section", etc.
        - AVOID words like 'listed', 'list', 'table', 'column', 'row', 'field', 'section', 'caption', 'metadata', etc., which implicitly or explicitly refer to the table or metadata.

1.4. **Naturalness**: The question should be phrased in a natural and conversational manner, as if it were being asked by a real data analyst or researcher.
    - Use simple, clear terminology. Avoid jargon, use layman-term descriptions instead.
    - The wording must be fluent and coherent, with proper grammar and syntax. It should read like a question that a human would naturally ask when analyzing data.

1.5. If the input `condition_col` is not "No condition column", the generated question MUST include a clear condition or constraint that references that column.
    - The condition should be derived from the values present in that column (inspect the column contents) and may be:
        - a filter on categorical values (e.g., "for [the object of question] with [condition_col] is 'Value'"),
        - a numeric comparison (e.g., "consider [the object of question] with [condition_col] > 1000", "regarding ... among in the top 10% of [condition_col]"),
        - a temporal or extrema-based condition (e.g., "in the year when [condition_col] reached its maximum", "for the month with the lowest [condition_col]"),
        - a ranking-based constraint (e.g., "among the top 3 entries by [condition_col]").

1.6. Answer:
    - The question must have a deterministic, single answer.
    - Ensure the question is answerable using only data present in the table.
    - Keywords of the question that will be used to determine the answer should be present in the table and not cover multiple values in the corresponding column.
        - E.g., "Existing" vs "Existing with temporary stands" should be treated as different keywords, and the question should be specific to one of them, such as "for the existing stadiums" or "for the stadiums with temporary stands".
    - For the ease of post evaluation, the answer must be a specific value (a **short** string, a number) or a small set of information (a list or an entity-value dictionary with no more than 3 elements or key-value pairs).
        - Example: {{'team_home': 'Poli Ejido', 'team_away': 'Espanyol'}}, "Espanyol", 83.6, ["Espanyol", "Poli Ejido"], etc.

1.7. **Excellent, Good and Bad Examples**:
    - EXCELLENT questions:
        - Based on the 2014 Pew Research Center survey data comparing the ethnic composition of Latter-day Saints in the U.S. to the general U.S. population in 2020, which ethnicity had the largest positive percentage point difference between its representation among Latter-day Saints and its representation in the general U.S. population?
        - Among the regional groupings for which both the International Monetary Fund and the World Bank provided data, which region showed the largest absolute difference between its 2025 IMF forecast and its 2024 World Bank estimate?
        - In the 2025-26 NCAA Division I women's basketball season, among the recorded upsets where a non-Division I team defeated a Division I opponent, which game featured the smallest point differential between the winner and the loser?
        - In the study of standard reduction potentials for aqueous iodine species, which chemical couple exhibits a potential of exactly 0 volts in basic conditions while simultaneously having the highest recorded potential among all such couples in acidic conditions?
        - Based on the 2022 cobalt production and reserves data reported by the USGS, what is the ratio of reserves to annual production of the group of all explicitly reported Asian countries (excluding aggregate categories like 'Other countries' or 'World total')?
    - GOOD questions:
        - What was the amount spent in millions of nominal dollars by the highest spending U.S Federal Department in the fiscal year of 1955?
        - Who holds the all-time record at the Grammys for the most wins in the album of the year category?
        - What is the current age of the oldest person to sail solo across the Pacific Ocean?
        - How many NBA players have scored 60 or more points in a regular season game since 2023?
        - Of the countries with a head of state assuming office prior to 2000, which are the five with the largest GDP per capita?
        - How many current and former Real Madrid players are ranked in the top 10 of the 2025 Forbes list of the world's highest-paid athletes?
    - BAD questions (and reason why each is bad):
        - Among the listed cities, which city has the highest population while being located in a state whose 2019 HDI is greater than 0.73? --> The phrase "Among the listed cities" is not clear enough to make the question fully self-contained.
        - Among the venues owned by AC Milan & Inter Milan for the 2026 Winter Olympics (as listed under a table in 'Milan cluster' section), which venue has the highest seating capacity? --> The note inside the parentheses explicitly refers to the table and section, which is not allowed.
        - Which venue in the Valtellina cluster has the highest total spectator capacity when the capacities of all its assigned events are summed together? --> Not fully self-contained.
        - What is the spectator capacity of the Verona Olympic Arena, the venue for the closing ceremony of the 2026 Winter Olympics? --> Too simple.


2. **SQL Implementation:**
    - Write a SQL query assuming the table is loaded as `df` and the engine is SQLite.
    - Ensure the query is syntactically correct and optimized for performance.
    - Always wrap column names with quotes, as some column names may contain spaces or special characters.
    - Ensure the query returns a single definitive value or a small result set.
    - The query must produce a non-empty result.

Now, generate a question and its SQL implementation based on the above instructions and the following input.

[[ ## table ## ]]
{table}

[[ ## column_name ## ]]
{column_name}

[[ ## metadata ## ]]
{metadata}
\end{rawprompt}
\noindent\rule{\linewidth}{0.5pt}

\textbf{Question quality checking}

\begin{rawprompt}[]
You are an expert in Q&A.

You are given a question, which is expected to be complex, self-contained, and natural.
Your task is to briefly discuss the question in the following criteria, then on a scale from 1 to 5 (1 is worst, 5 is best), rate the question on each criterion:
- Complexity: The question involves multi-hop reasoning and information aggregation.
- Self-containedness: The question provides all necessary parameters, definitions, and background information within itself, requiring no external links, data, references, or prior knowledge to be **understood**. Using the question, people can expect to search for all necessary information to answer it from the Internet. 
- Naturalness: The question is as natural as being asked by real human.

Now, evaluate the question based on the above criteria and provide a brief explanation for each rating.

[[ ## question ## ]]
{question}
\end{rawprompt}
\noindent\rule{\linewidth}{0.5pt}

\textbf{Python solution generation}

\begin{rawprompt}[]
You are an expert in table question answering and problem solving with Python.

You are given a table in Markdown format, some metadata about the table, and a complex question about the table.
Your task is to generate a Python implementation to answer the question. You must name the final answer variable `final_answer_python`.

The answer is expected to be a specific value (a short string, a number) or a small set of information (a list or an entity-value dictionary with no more than 3 elements or key-value pairs). Please only pack the final answer in `final_answer_python` variable, and avoid putting any narration in `final_answer_python` (e.g., "the answer is ...", "the result is ...", etc.).

**Guidelines:**
  - Use `pandas` to solve the problem, **assuming the table is already loaded into a DataFrame named `df`**.
  - Skip the data loading step.
  - **Break complex operations into small steps.** Each step should compute one logical operation.
  - Assign each intermediate result to a named variable (e.g., `step1_filtered`, `step2_grouped`).
  - **Do not reuse the same variable name for different steps**.
  - If the change is in a row or column of an existing variable, create a new variable to store the updated version (e.g., `step1_filtered` -> `step2_filtered`).
  - Each step should have a unique variable name and should reflect the meaning of the data in the variable.
  - Display the intermediate results and final result by printing them.
  - Comment your code to explain each step clearly.
  - **Use the provided context about the table to narrate your solution**.
  - Name the final answer variable `final_answer_python`.

Now, generate a Python solution based on the above instructions and the following input.

[[ ## table ## ]]
{table}

[[ ## metadata ## ]]
{metadata}

[[ ## question ## ]]
{question}
\end{rawprompt}
\noindent\rule{\linewidth}{0.5pt}

\textbf{Dual-path answer verification and \dataset-Bench's LLM grader}

\begin{rawprompt}[]
You are an expert in data analysis.

You are given two results (tables, values, or dictionaries) from answering the same question about a dataset.
Your task is to compare if the content/meaning of the two results are essentially the same.

**Metadata is provided** to help you understand what the question was asking and what the results represent.

**Comparison Guidelines:**
- The two results may have different formatting (e.g., table vs. dictionary, different column order, different indexing)
- They may have slight numeric differences due to floating-point precision (e.g., 83.6 vs 83.60000000000001)
- As long as the semantic content of results regarding the question are the same, consider them equivalent
- Focus on whether both results answer the question correctly

**Examples of equivalent results:**
- "Espanyol" vs. {"team_home": "Poli Ejido", "team_away": "Espanyol", "team_advances": "Espanyol"} if the question is "Which team advances to the next round?" and both results indicate Espanyol advances
- {"party": "Democratic", "avg": 83.6} vs {"party": "Democratic", "avg": 83.60000000000001}
- A single-row table vs a dictionary with the same key-value pairs
- 248Cm vs. {}^{248}Cm vs. 248Cm (i.e., {}^{248}Cm): all of them represent the same chemical element isotope, just different in the way of writing.

Give a binary answer (True or False) and a short explanation for your judgment.

Now compare the two results based on the above instructions and the following input.

[[ ## result1 ## ]]
{result1}

[[ ## result2 ## ]]
{result2}
\end{rawprompt}
\noindent\rule{\linewidth}{0.5pt}

\textbf{Reasoning trace generation (Back Translation)}

\begin{rawprompt}[]
You are given a question, a set of context documents/chunks, and the correct answer.
Your task is to generate a detailed, step-by-step reasoning trace that explains how to arrive at the answer by analyzing the context.

## Input Format

### **Context:**
{context}

### **Question:**
{question}

### **Correct Answer:**
{answer}

## Output Format Requirements

Generate a natural-language reasoning trace following these structural patterns:

### 1. Step Numbering
- Use `* Step X:` format for major steps
- Steps should progress logically from understanding the problem to arriving at the answer

### 2. Planning Steps (Initial Steps)
- Start by analyzing what the question is asking
- Identify what information needs to be extracted from the context
- Determine the search strategy (e.g., "find all candidates for each criterion")
- Example: "The question asks about X that fits Y criteria, so we need to find all candidates..."

### 3. Extraction & Navigation Steps
- Systematically review relevant documents/chunks
- Document your search process (e.g., "let's search for keyword X: found in documents A, B, C")
- For each document examined, state whether it contains relevant information
- Use phrases like:
  - "Let's find..."
  - "Document X says..."
  - "It fits/does not fit because..."
  - "Let's check..."
  - "Move to document..."

### 4. Iterative Filtering
- For multi-criteria questions, evaluate candidates against each criterion
- Narrow down the candidate list as you verify criteria
- Explicitly state when a candidate is eliminated and why
- State when a candidate passes a criterion

### 5. Evidence Citation
- Always cite specific documents/chunks when stating facts
- Quote relevant passages when they directly support conclusions
- Use formats like:
  - "Document X states: '...'"
  - "According to document X,..."

### 6. Self-Verification
- Include steps that cross-check findings
- Verify consistency across multiple sources when available
- Address potential contradictions explicitly

### 7. Reasoning Steps
- Perform virtual operations (filtering, sorting, comparing) explicitly
- When calculating, show the calculation
- When comparing, state the comparison clearly
- Build confidence through multiple verified facts before final answer

### 8. Conclusion
- The final step should clearly state the answer
- Briefly summarize the key evidence that leads to it

## Style Guidelines

1. **Natural and exploratory**: Write as if discovering the answer in real-time
2. **Self-contained**: Each step should be understandable without re-reading previous steps
3. **Critical evaluation**: Don't accept information at face value; evaluate relevance
4. **Document-focused**: Ground all conclusions in the provided context
5. **Conversational but precise and concise**: Use natural language while maintaining factual accuracy and efficiency

## Example Structure

```
* Step 1: [Understanding the question and planning strategy]
* Step 2: [Search for candidates matching criterion A - examining documents]
* Step 3: [Search for candidates matching criterion B - narrowing down]
* Step 4: [Cross-referencing and verification]
* Step 5: [Conclusion with answer]
```

## Important Notes

- The reasoning trace should be answer-agnostic in style (don't reveal you know the answer upfront)
- Show the discovery process as it would happen when solving the problem
- Include dead-ends and corrections when relevant (e.g., "Wait, this document is from 2018, so it may be outdated")
- If multiple interpretations exist, explore them before concluding

Now generate the reasoning trace for the given input.

\end{rawprompt}
\noindent\rule{\linewidth}{0.5pt}

\clearpage
\begin{figure*}[t]
\centering
\includegraphics[width=\textwidth]{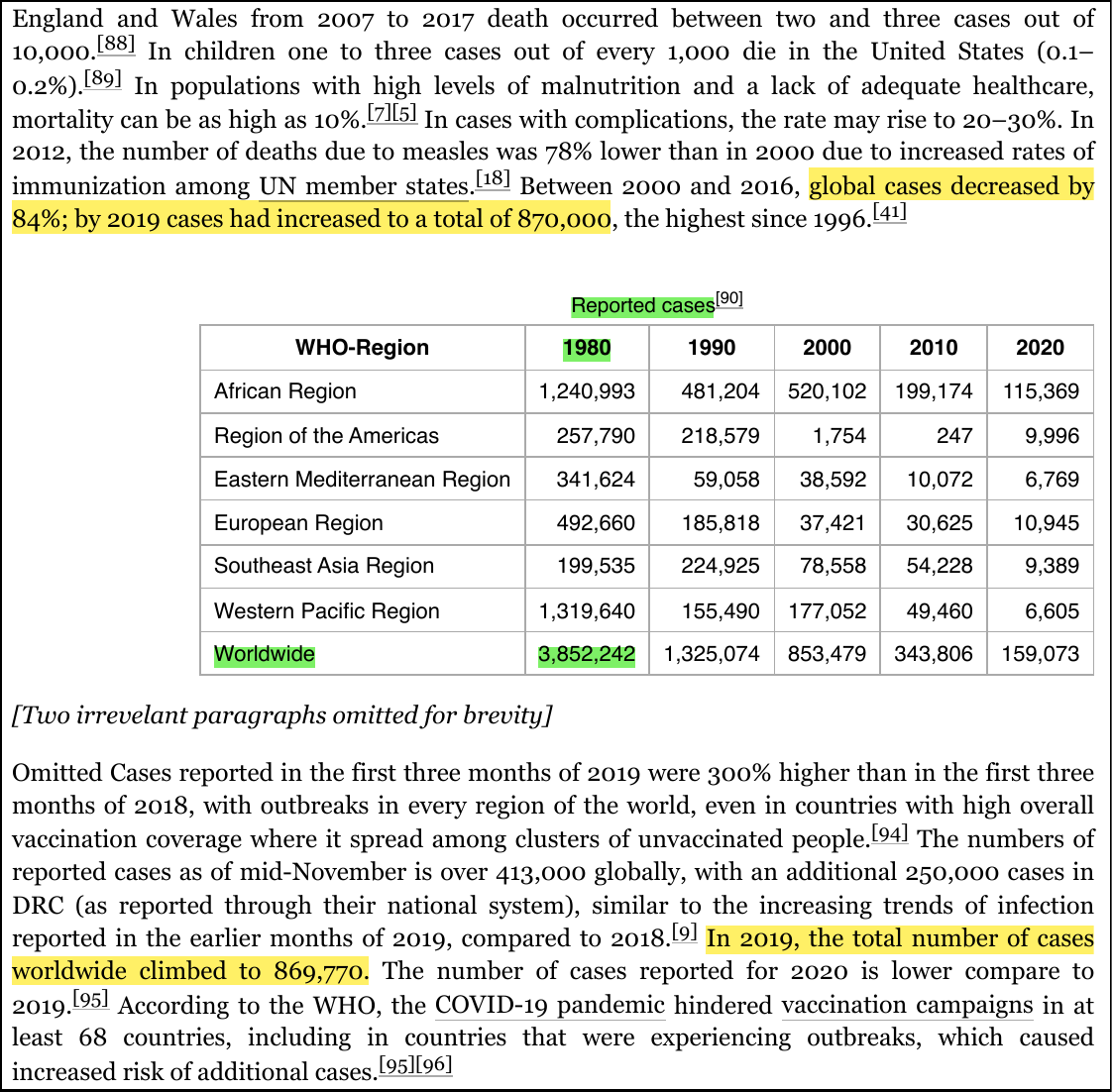}
\caption{The evidence as a part of the context of the question \textit{"According to the WHO, in which year were the highest number of measles cases recorded worldwide?"} in the \textbf{LongSeal} benchmark. Retrieved from the Wikipedia page titled "Measles" on May 20, 2026. Given the context, the answer of the question is 1980. \colorbox{green}{Green-highlighted} text notes the evidence that leads to the correct answer, while the \colorbox{yellow}{yellow-highlighted} text notes the evidence that may lead to a misunderstanding and thus an incorrect answer.}
\label{fig:context-of-good-sample}
\end{figure*}

\end{document}